\ifdefined\XeTeXversion\else\pdfoutput=1\fi
\PassOptionsToPackage{table}{xcolor}
\documentclass[10pt,logo,onecolumn,copyright]{nv}

\usepackage{graphicx}
\usepackage{xcolor}
\usepackage[utf8]{inputenc}
\usepackage[T1]{fontenc}
\usepackage{times}
\usepackage{amsmath}
\usepackage{amssymb}
\usepackage{mathtools}
\usepackage{booktabs}
\usepackage{multirow}
\usepackage{tabularx}
\usepackage{subcaption}
\usepackage{float}
\usepackage{url}
\usepackage[nameinlink]{cleveref}
\usepackage[square,sort,comma,numbers]{natbib}
\usepackage{xspace}
\usepackage{url}
\usepackage{booktabs}
\usepackage{graphicx}
\usepackage{wrapfig}
\usepackage{needspace}
\usepackage{arydshln}

\definecolor{nvidiagreen}{HTML}{76B900}
\definecolor{nvgreen}{HTML}{D3E9AD}
\definecolor{nvgreenlight}{HTML}{EAF5D8}
\definecolor{nvgray}{HTML}{5B5E5F}
\hypersetup{
  colorlinks=true,
  citecolor=nvidiagreen,
  linkcolor=nvidiagreen,
  urlcolor=nvidiagreen
}

\title{Long-WAM: Scaling the Context of World-Action Models}

\runningtitle{Long-WAM: Scaling the Context of World-Action Models}

\author{
\parbox{\linewidth}{
\centering
\vspace{-5pt}
{\fontsize{9pt}{16pt}\selectfont\textbf{Wei Huang\textsuperscript{1\thinspace3\thinspace*}, ~ Bohan Zhang\textsuperscript{2\thinspace*}, ~ Chenzhi Liu\textsuperscript{3\thinspace}, ~ Isabella Liu\textsuperscript{1\thinspace4}, ~ Shuai Yang\textsuperscript{1\thinspace}, ~ Weian Mao\textsuperscript{1\thinspace}}}
\\
\vspace{0.35em}
{\fontsize{9pt}{16pt}\selectfont\textbf{Luozhou Wang\textsuperscript{1}, ~ Yicheng Xiao\textsuperscript{3}, ~ Weifeng Lin\textsuperscript{1}, ~ Qixin Hu, ~ Bryan Chu\textsuperscript{1}, ~ Sifei Liu\textsuperscript{1}}}
\\
\vspace{0.5em}
{\fontsize{9pt}{16pt}\selectfont\textbf{Linxi “Jim” Fan\textsuperscript{1}, ~ Xiaojuan Qi\textsuperscript{3}, ~ Song Han\textsuperscript{1\thinspace2}, ~ Yukang Chen\textsuperscript{1}}}
\\
\vspace{2.2mm}
{\fontsize{8.7pt}{11pt}\selectfont \textsuperscript{1}NVIDIA \quad \textsuperscript{2}MIT \quad \textsuperscript{3}HKU \quad \textsuperscript{4}UCSD}
\\
\vspace{5pt}
{\fontsize{8.2pt}{10pt}\selectfont \textsuperscript{*} Equal contribution. }
\vspace{10pt}
\\
{\fontsize{9pt}{11pt}\selectfont
\href{https://github.com/NVlabs/LongLive/tree/main/Long-WAM}{\raisebox{-0.15em}{\includegraphics[height=1em]{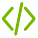}}\hspace{0.35em}\textbf{Code}}
\hspace{1.8em}
\href{https://nvlabs.github.io/LongLive/Long-WAM/}{\raisebox{-0.15em}{\includegraphics[height=1em]{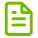}}\hspace{0.35em}\textbf{Project Page}}}
\\
\vspace{-10pt}
}}

\begin{abstract}\small
\textbf{Abstract: }Real-time robot control demands enough visual history to infer motion and
task progress, but processing that history can delay action.
We present \textbf{Long-WAM}, a model--system framework for scaling the context
of causal world-action models under real-time control constraints.
Our central finding is that access to history is not the same as using it:
longer histories pay off far more when the video foundation is pretrained
autoregressively (AR).
We first learn causal prediction from robot and egocentric videos without
action labels, then preserve this history-to-future structure during
world-action adaptation.
On RoboCasa GR-1, increasing context from 0.0 to 19.2 seconds raises success
from 63.3\% to 78.7\%, whereas a bidirectionally pretrained initialization shows
no net gain; robot-domain AR pretraining further raises peak success on GR-1
and LIBERO-Long. Long-WAM also achieves the best results among compared methods
on LIBERO-Long, RoboTwin~2.0, and DOMINO. Streaming observation encoding,
asynchronous execution, and hardware-specific acceleration enable deployment
on RTX~5090, DGX Spark, and Jetson AGX Thor without dropping future prediction;
on RTX~5090, each action chunk, including future-video latent prediction, takes
107.4\,ms. Real-time deployment on Unitree G1 and YAM supports dynamic and
long-horizon manipulation, including 95\% success on dynamic cup stacking,
where $\pi_{0.5}$ and Fast-WAM succeed in none of 20 trials.
As a memory-informed executor, Long-WAM also complements higher-level planning
in composite tasks.
\end{abstract}

\begin{document}

\maketitle


\begin{figure}[h]
\centerline{\includegraphics[width=1\textwidth]{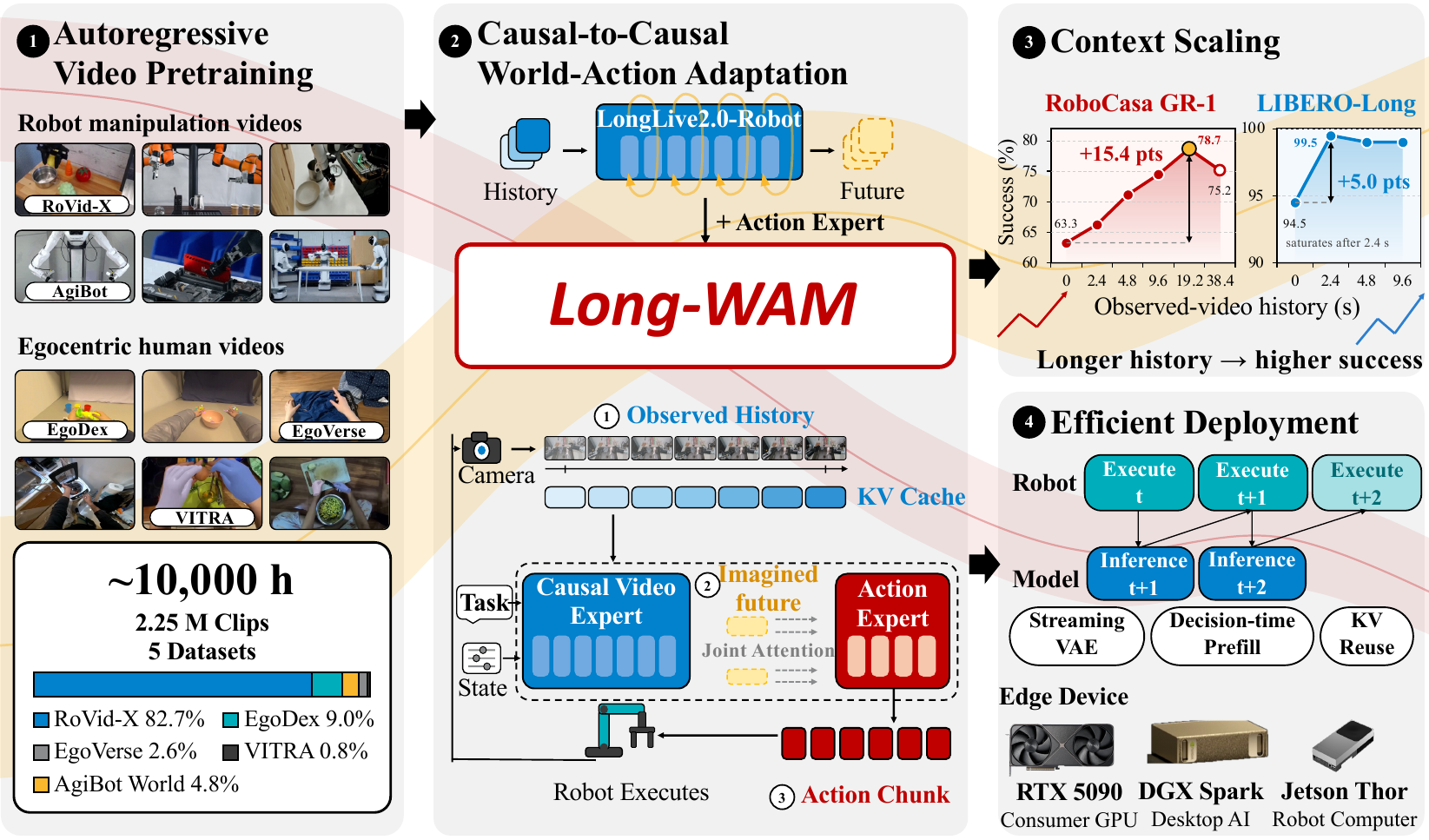}}
\vspace{-2pt}
\caption{\textbf{Overview of Long-WAM.}
\textbf{(1) AR video pretraining.} LongLive2.0-Robot learns predictive
dynamics from approximately 10,000 window-equivalent hours of robot and
egocentric videos.
\textbf{(2) Causal-to-causal adaptation.} We preserve causal video dependencies
while conditioning action-chunk denoising on observed history.
\textbf{(3) Context scaling.} Relative to current-observation-only control,
success increases from 63.3\% to 78.7\% on RoboCasa GR-1 and from 94.5\% to
99.5\% on LIBERO-Long, peaking at 19.2 and 2.4 seconds of history, respectively.
\textbf{(4) Efficient deployment.} Asynchronous execution, streaming VAE
encoding, decision-time prefill, within-call KV reuse, and hardware-specific
acceleration enable real-time control on RTX~5090, DGX Spark, and Jetson AGX Thor.}
\label{fig:overview}
\end{figure}




\Needspace{12\baselineskip}
\clearpage

\section{Introduction}

Fast robot control requires understanding how the scene is changing.
A single image can reveal an object's position but leave its motion and
interaction progress ambiguous. World-action models (WAMs) bring video
prediction to closed-loop control \citep{bi2025motus,ye2026dreamzero,li2026lingbotva},
making visual history a natural resource for action. Yet processing more
history can delay the response it is meant to improve. We therefore ask:
\emph{how does WAM control scale with visual context, and how can those
gains be retained under real-time constraints?}

Recent WAMs retain history through causal caches and persistent or selected memories
\citep{yang2026memorywam,sun2026himemwam,wang2026dimwam,su2026worldscape2}.
Yet \emph{access to history is distinct from learning to predict from it}.
DreamZero
\citep{ye2026dreamzero} and LingBot-VA \citep{li2026lingbotva} adapt
bidirectionally pretrained video generators to causal video--action prediction,
whereas LingBot-VA 2.0 \citep{zhang2026lingbotva2} jointly pretrains causal
video and learned latent actions. We investigate a complementary route:
first learn autoregressive (AR) video prediction,
then adapt to actions while preserving its history-to-future structure. Appendix~\ref{sec:related} discusses related work.

We introduce \textbf{Long-WAM} (Figure~\ref{fig:overview}), a model--system
framework for context scaling in real-time robot control. Our central finding
is that \emph{the value of longer context depends on how the video foundation
is pretrained} (Figure~\ref{fig:memory-scaling}). On GR-1, both AR-pretrained
foundations gain from extending context from 0.0 to 19.2 seconds, whereas a
bidirectional initialization does not; the robot-domain AR model's lead
over it grows from 3.3 points without history to 17.1 points at 19.2 seconds.
AR pretraining already learns the causal temporal factorization that WAM
adaptation retains, giving the model a basis for using history rather than
merely accessing it. Our LongLive2.0-Robot foundation learns from roughly
10,000 window-equivalent hours of robot and egocentric video; we preserve its
causal structure during action adaptation, and robot-domain pretraining further
raises peak success on both LIBERO-Long and GR-1.

Longer context helps only if the controller still responds in time, so we
co-design asynchronous execution and edge acceleration while retaining future
prediction. Predictive conditioning from the AR video foundation supports
smooth action handoffs without blending or prefix guidance; streaming VAE
encodes incoming observations while the robot executes, reducing post-trigger
computation; and NVFP4 quantization with device-specific kernel tuning
accelerates the video--action pipeline on RTX~5090, DGX Spark, and Jetson AGX
Thor. With four denoising steps per expert, the optimized pipeline takes
107.4\,ms per action chunk on RTX~5090, including the full observation VAE
computation, a $3.3\times$ speedup over BF16 eager execution
(Section~\ref{sec:infrastructure}, Table~\ref{tab:efficiency-cumulative-ablation}).

On RoboCasa \citep{nasiriany2024robocasa} GR-1 Tabletop
\citep{nvidiagear2025gr1tabletop}, increasing the history window from 0.0 to
19.2 seconds raises success from 63.3\% to 78.7\% (+15.4 points).
On LIBERO-Long \citep{liu2023libero}, 2.4 seconds of history improves
success from 94.5\% to 99.5\%. Long-WAM also achieves 94.4\% average success
on RoboTwin~2.0 \citep{chen2025robotwin2} and 34.9\% success on DOMINO
\citep{fang2026domino}, the highest among compared methods.
On a Unitree G1, Long-WAM grasps cups from a conveyor moving at 7.5~cm/s in
90\% of trials and stacks moving cups in 95\%, settings in which neither
$\pi_{0.5}$ \citep{physicalintelligence2025pi05} nor Fast-WAM
\citep{yuan2026fastwam} succeeds once. On YAM, it completes tasks lasting over
40 seconds on average with 81.7\% success, extending the evaluation from rapid
interception to sustained execution. Long-WAM also serves as the executor of a
hierarchical system on RoboCasa365 \citep{nasiriany2026robocasa365}: pairing
its unchanged checkpoint with GPT-6 Astra raises overall success from 31.4\%
to 54.4\%, versus 25.2\% for the planner alone. Planner augmentation yields a
larger gain for Long-WAM (+23.0 points) than for $\pi_{0.5}$ (+13.6;
Section~\ref{app:robocasa365-planning}), suggesting that stronger execution
lets planning pay off more. Together, these results suggest that context is a
resource for generative control whose value depends on both predictive
pretraining and timely execution.

\section{Related Work}
\label{sec:related}

\paragraph{World-action modeling from video generators.}
WAMs connect visual dynamics to executable actions.  Motus \citep{bi2025motus}
combines pretrained video, vision--language, and action experts; DreamZero
\citep{ye2026dreamzero} and LingBot-VA \citep{li2026lingbotva} adapt
bidirectional video backbones to causal control.  Their video pretraining
supplies visual priors, while causal prediction and action coupling are learned
downstream.  LingBot-VA 2.0 \citep{zhang2026lingbotva2} instead jointly pretrains
causal video and learned latent actions in a shared representation.
EVA \citep{wang2026eva} aligns video generation with executable actions through
inverse-dynamics rewards, and $\omega$-EVA \citep{sun2026omegaeva} evaluates
latent action consequences.  Long-WAM develops a staged route: action-free
robot-video AR pretraining followed by action adaptation under the same causal
ordering.  The action model thus inherits a visual dynamics prior explicitly
trained for history-to-future prediction, separating predictive pretraining
from learning an embodiment's action representation.

\paragraph{Memory and context scaling.}
Cache capacity, physical history duration, and the cost of rebuilding visual
context are distinct quantities. Causal caches, compressed histories, and
event retrieval provide complementary memory interfaces
\citep{ye2026dreamzero,li2026lingbotva,yang2026memorywam,
sun2026himemwam,wang2026dimwam,su2026worldscape2}. Echo-Memory \citep{king2026echomemory} studies memory for
camera-conditioned video generation; RoboTTT \citep{jiang2026robottt} studies
context scaling in robot policies; WAM-TTT \citep{feng2026wamttt} adapts memory
from human video to steer a frozen WAM.  Long-WAM studies the robot's own
observed interaction history: we vary its temporal extent across trained
causal WAM variants and examine closed-loop success and online computation.
Our focus is how predictive pretraining shapes the benefit of longer context,
complementing work on memory compression, retrieval, and adaptation.

\paragraph{Efficient and deployable generative control.}
Large video backbones make online WAM control expensive.  Prior systems reduce
this cost through caching, few-step generation, pipelined execution, action-only
decoding, or hierarchical update rates
\citep{ye2026dreamzero,li2026lingbotva,yuan2026fastwam,cai2026ahawam}.
LongLive-2.0 further develops parallel AR training, compressed KV caches,
low-precision execution, and asynchronous VAE decoding
\citep{chen2026longlive2}.  Complementary work on compilation, kernel
orchestration, and video-DiT quantization reduces inference cost
\citep{ansel2024pytorch2,hu2024korch,zhao2024viditq}.
Asynchronous deployment additionally requires consistency between consecutive
action chunks.  Existing methods constrain these transitions
through inference-time guidance, prefix-conditioned training, or denoising-time
blending \citep{black2025rtc,black2025trainingrtc,lin2026holobrain}.
An empirical WAM study highlights the importance of temporal alignment and the
precision--smoothness trade-offs of transition strategies
\citep{motubrainteam2026realtime}.  Long-WAM exhibits robust
continuity under direct asynchronous switching while retaining explicit future
visual conditioning.  We preserve this imagine-then-act path and address its
sequential cost through streaming observation encoding and device-specific
acceleration, targeting continuity and speed together.

\begin{figure*}[!t]
\centering
\includegraphics[width=1.0\textwidth]{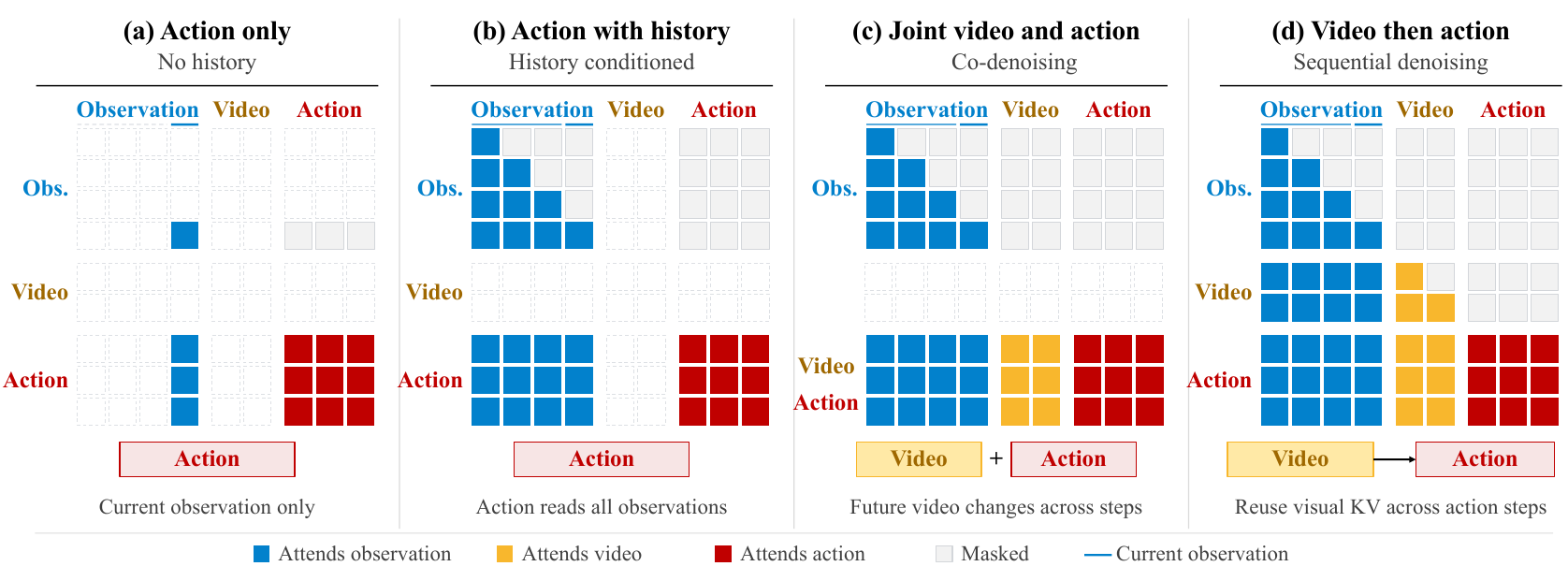}
\vspace{-10pt}
  \caption{\textbf{Attention patterns and video--action inference strategies.}
  (a) Action generation from the current observation only;
  (b) history-conditioned action generation without future prediction;
  (c) joint video--action co-denoising (CoD);
  (d) video prediction followed by action denoising (IDM).
  IDM conditions actions on both observed history and predicted future latents,
  reusing visual KV across action-denoising steps.}
\label{fig:wam-attention}
\end{figure*}

\section{Scaling the Context of WAMs from AR Video Generation}
\label{sec:method}
\vspace{-5pt}

Long-WAM connects context scaling to the ability to predict physical evolution
from history. We first learn robot motion and interaction dynamics through
long-sequence AR video pretraining, then transfer this predictive foundation
to action generation while preserving its causal temporal structure.
The resulting policy conditions actions on both observed history and
anticipated futures, making prediction an intermediate representation for
control. We then study how the benefit of additional history depends on the
video initialization, distinguishing access to a longer context from the
learned ability to exploit it.

\subsection{LongLive2.0-Robot: Robot-Domain AR Video Pretraining}
\label{sec:pretraining}

\paragraph{Data and initialization.}
LongLive2.0-Robot continues training from LongLive-2.0's 16-second AR
checkpoint \citep{chen2026longlive2} on approximately 10,000 window-equivalent
hours from RoVid-X \citep{deng2026rovidx}, AgiBot World
\citep{agibotworld2025}, EgoDex \citep{hoque2026egodex}, EgoVerse
\citep{punamiya2026egoverse}, and VITRA \citep{li2025vitra}.
Because supervision is video-only, the model can learn from multiple embodiments without requiring a shared action space. Appendix~\ref{app:data} details the data accounting.

\paragraph{Teacher-forcing training.}
We pretrain on robot-video sequences up to 30 seconds long, exposing the
model to extended motion and interaction histories. To support this temporal
span, we adopt LongLive-2.0's sequence-parallel AR training
\citep{chen2026longlive2}, which shards long sequences across GPUs.
Following the teacher-forcing formulation for AR video generation \citep{zhou2025taming}, block-causal attention supervises every noisy chunk from its ground-truth prefix in parallel, while the conditioning image stays clean and outside the loss.
We retain LongLive-2.0's error recycling, derived from SVI \citep{li2026stable}.
For a clean target chunk $z_i$, let $\bar z_i$, $\bar\epsilon_i$, and $h_{<i}$
denote the target latent, sampled Gaussian noise, and preceding ground-truth
context after optional buffered-error perturbations. The noisy input and
teacher-forcing objective are
\begin{equation}
x_i^{\sigma_i}=(1-\sigma_i)\bar z_i+\sigma_i\bar\epsilon_i,
\qquad \sigma_i\in[0,1],
\label{eq:video-noising}
\end{equation}
\begin{equation}
\mathcal L_{\mathrm{TF\text{-}AR}}=\mathbb E\!\left[w(\sigma_i)\left\|v_\theta(x_i^{\sigma_i},\sigma_i\mid h_{<i},c)-(\bar\epsilon_i-z_i)\right\|_2^2\right].
\label{eq:ar-video}
\end{equation}
Here $\sigma_i$ is the sampled noise level, $v_\theta$ the video velocity
predictor, $c$ the language condition, and $w$ the scheduler weight.
The recovery target retains the original $z_i$, teaching correction of
rollout-like errors.
Given one image and a language prompt, the model predicts coherent robot
motion and object interactions (Appendix~\ref{app:video-predictions}),
providing a predictive prior for action adaptation.

\subsection{Causal-to-Causal World-Action Adaptation}
\label{sec:adaptation}

\paragraph{Causal coupling.}
Following DreamZero \citep{ye2026dreamzero} and LingBot-VA \citep{li2026lingbotva}, we couple video and action experts through an asymmetric interface. Video queries read only their own and
earlier visual blocks, never action tokens; action queries read observed
history, partially denoised futures, and the entire noisy action chunk.
This preserves pretrained causal visual dependencies while grounding actions
in both past and anticipated interaction.
Crucially, the history-to-future ordering is learned during video pretraining,
not introduced only at action adaptation.
At a decision made at control step $t$, let $Z_t^-$ denote observed video latents, $q_t$ the robot
state, $c$ the language instruction, and $\mathbf A_t=a_{t:t+H-1}$ an $H$-step action
chunk. The video expert $\theta$ predicts $K_v$ future latent steps
$\widetilde Z_t^+$ to noise level $\sigma_\star\in(0,1)$, then supplies their
joint visual cache $\mathcal K_t$ to action expert $\psi$:
\begin{equation}
\begin{aligned}
\widetilde Z_t^+&=\operatorname{Rollout}_{\theta,\sigma_\star}(\epsilon^v\mid Z_t^-,c,q_t),
&\mathcal K_t&=\operatorname{Prefill}_{\theta}(Z_t^-,\widetilde Z_t^+;c,q_t,\sigma_\star),\\
\widehat{\mathbf A}_t&=\operatorname{Denoise}_{\psi}(\epsilon^a\mid\mathcal K_t,c,q_t).
\end{aligned}
\label{eq:wam}
\end{equation}
Here $\epsilon^v$ and $\epsilon^a$ are independent Gaussian noise, $\mathcal K_t$ contains layer-wise video keys and values, and observed latents remain clean throughout. We refer to this predict-then-act mode as inverse dynamics modeling (IDM, Figure~\ref{fig:wam-attention}).

\paragraph{Training and inference.}
Training uses two passes: video flow matching conditioned on clean history,
then action flow matching conditioned on history and a forward-noised
ground-truth future, using $\sigma_\star=0.9$.
The second pass detaches the visual cache, so action loss updates only the
action expert and proprioceptive adapter. Both branches predict
noise-minus-data velocity, with weighted objective
$\mathcal L=\lambda_v\mathcal L_{\mathrm{video}}+\lambda_a\mathcal L_{\mathrm{action}}$,
where $\lambda_v$ and $\lambda_a$ balance the losses.
At inference, the four video steps (V4) stop at $\sigma_\star=0.9$ rather than at a clean video, and future latents are never decoded to pixels; their cache is reused throughout action denoising. Appendix~\ref{app:training} specifies the training surrogate,
masking, and cache scope.

\begin{figure*}[!t]
\centering
\includegraphics[width=1.0\textwidth]{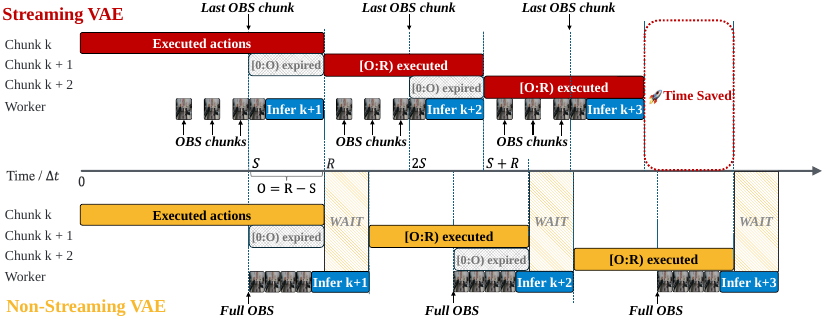}
\vspace{-10pt}
\caption{\textbf{Asynchronous model--robot execution with streaming VAE.}
Both schedules use the same nominal trigger stride $S$ and overlap $O=R-S$.
Streaming VAE (top) encodes observation (OBS) chunks as they arrive and meets
the handoff deadline $T_{\mathrm{ready}}\leq O\Delta t$.
Full-window encoding (bottom) delays handoffs and subsequent triggers, accumulating robot idle time.
Time is shown in units of $\Delta t$.
}
\label{fig:infra-async}
\end{figure*}

\subsection{Context Scaling of World-Action Models}
\label{sec:context}

We scale the duration of real observations in the causal prefix, keeping the
visual forecast and action horizon fixed within each benchmark. This tests
whether additional past evidence improves the same near-term control decision,
rather than changing how far the model predicts or acts. Each window
uses a separately trained model evaluated at its training context length;
zero history retains the current observation. Missing early-episode history repeats the initial
frame.

We compare Wan2.2 (bidirectional), LongLive-2.0 (AR), and LongLive2.0-Robot
(robot-domain AR) initializations under causal WAM adaptation. Here,
bidirectional describes pretraining, not the adapted policy's temporal mask.
The question is whether longer context yields greater control benefits when
causal prediction is learned before action adaptation. We sweep
0--38.4 seconds on RoboCasa GR-1 Tabletop
\citep{nasiriany2024robocasa,nvidiagear2025gr1tabletop}, with a complementary
context study on LIBERO-Long \citep{liu2023libero}
(Section~\ref{sec:scaling-experiments}). Longer prefixes also increase prefill,
attention, and cache costs. We therefore treat context as an execution resource
whose value depends on both predictive benefit and response time; the following
infrastructure section addresses this deployment cost.

\section{Long-WAM Infrastructure: Real-Time Edge Deployment}
\label{sec:infrastructure}

\begin{figure*}[t]
\centering
\includegraphics[width=1.0\textwidth]{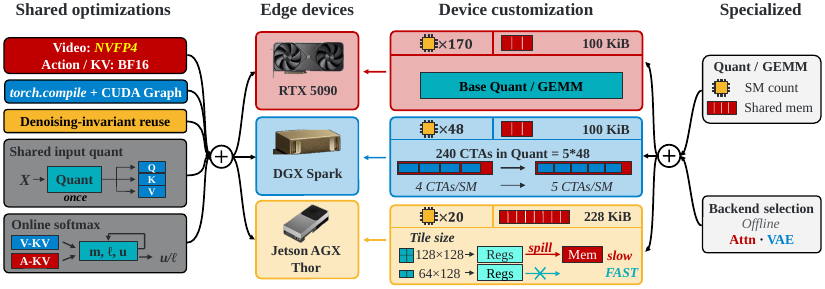}
\caption{\textbf{Optimizations for efficient edge deployment.}
Shared optimizations and device-specific tuning accelerate edge inference. ``Base'' denotes the Quant/GEMM
implementation; shape-specific dispatch and autotuning remain enabled.
}
\label{fig:infra-acceleration}
\end{figure*}

Real-time deployment must accommodate both historical observations and
future prediction within the time available to prepare the next action chunk.
We address this constraint by co-designing asynchronous execution and edge
acceleration (Figures~\ref{fig:infra-async} and~\ref{fig:infra-acceleration}).
Asynchronous scheduling overlaps model inference with robot execution,
while streaming causal VAE encoding moves prefix observation encoding ahead
of the inference trigger. Shared optimizations, including NVFP4 quantization
and within-call KV reuse, combine with device-specific kernel tuning to
accelerate the video--action pipeline on RTX~5090, DGX Spark, and Jetson AGX
Thor. Together, these designs target timely action handoffs while retaining
the predictive conditioning that supports history-aware control.

\subsection{Asynchronous Execution}
\label{sec:async-execution}

\paragraph{Pure asynchronous execution suffices.}
We overlap inference with robot execution
without blending or prefix guidance. At decision $k$ (control step $t_k$),
the predicted chunk is $\widehat{\mathbf A}_{t_k}\in\mathbb R^{H\times d}$,
where $d$ is the action dimension (Section~\ref{sec:adaptation}).
Only the first $R$ steps are eligible for execution; inference has nominal stride $S$
($R/2\leq S<R\leq H$), leaving an overlap of $O=R-S$ steps.
At the handoff, the controller discards the elapsed prefix and executes the
new chunk's aligned suffix, waiting if it arrives late
(Figure~\ref{fig:infra-async}).

Pure asynchronous execution can disrupt action continuity
\citep{motubrainteam2026realtime}, motivating inference-time guidance
\citep{black2025rtc} and training-time action conditioning
\citep{black2025trainingrtc}. Long-WAM requires neither in our experiments.
Conditioned on temporally continuous LongLive2.0-Robot forecasts, consecutive
chunks tend to agree in their overlap: on RoboTwin~2.0
\citep{chen2025robotwin2}, Long-WAM nearly keeps its synchronous success under
asynchrony, whereas Fast-WAM \citep{yuan2026fastwam} and LingBot-VA
\citep{li2026lingbotva} lose 15.4 and 45.5 percentage points (Table~\ref{tab:async-continuity}).

\paragraph{Streaming VAE.}
Reducing overlap from 12 to 8 control steps lowers overlap RMSE about fourfold
and jerk nearly threefold (Appendix~\ref{app:async-overlap}). At fixed $R$,
shorter overlap requires later inference triggers and a tighter handoff deadline:
\begin{equation}
T_{\mathrm{ready}}\leq O\Delta t=(R-S)\Delta t,
\label{eq:async-deadline}
\end{equation}
where $\Delta t$ is the control interval and $T_{\mathrm{ready}}$ includes
transfer, queueing, and computation from the time the trigger observation becomes available.

Streaming causal VAE encoding processes incoming frame groups. After the trigger,
we encode the remaining frames, concatenate features, and project and normalize the latents. At the same $S$ and $O$,
it avoids full-window encoding's repeated waits and accumulated idle time
(Figure~\ref{fig:infra-async}), allowing later triggers and shorter overlap
while retaining the latest observation.

\subsection{Efficient Edge Deployment}
\label{sec:edge-acceleration}

Local inference avoids network delays, but video-first prediction is costly
on limited onboard compute. We accelerate Long-WAM on NVIDIA GeForce RTX~5090,
DGX Spark, and Jetson AGX Thor to support short-overlap execution
(Equation~\ref{eq:async-deadline}) while preserving visual imagination.
Shared optimizations reduce common computation and data-movement costs;
device-specific tuning addresses platform constraints, balancing portability and
hardware efficiency.

\paragraph{Shared optimizations.}
Across devices (Figure~\ref{fig:infra-acceleration}), video-expert linear layers
use W4A4 NVFP4 (four-bit weights and activations) during generation and key--value (KV) prefill \citep{nvidia2024blackwell,chen2026longlive2},
while action compute and KV storage remain BF16.
Action quantization offers limited latency savings and may compromise control
precision, so we retain BF16.
NVFP4 reduces compute and weight/activation storage, but adds small quantization
and scaling kernels whose launch overhead can offset the compute savings.
When NVFP4 is combined with CUDA Graph replay \citep{gray2019cudagraphs} and
PyTorch compilation \citep{ansel2024pytorch2}, launch overhead is reduced and
eligible operations are fused.
We also reuse denoising-invariant text/state KV, observed-video KV, and FP32 RoPE
tables \citep{su2021roformer} within each inference call to reduce redundant computation;
streaming-VAE operations that update state remain outside CUDA Graph capture.

Shared input quantization quantizes the shared input to Q/K/V projections once
and reuses the quantized activations and scales across three separate GEMMs.
We combine attention over separate video and action KV buffers using online softmax
\citep{milakov2018softmax} with a shared normalization, avoiding buffer concatenation.
For each query, let $(m_j,\ell_j,u_j)$ denote the maximum attention logit,
the sum of exponentials shifted by $m_j$, and their value-weighted sum for
segment $j\in\{v,a\}$; the merged output is
\[
m=\max(m_v,m_a),\qquad
\operatorname{Attn}=\frac{e^{m_v-m}u_v+e^{m_a-m}u_a}
{e^{m_v-m}\ell_v+e^{m_a-m}\ell_a}.
\]
This preserves joint attention without a concatenated KV buffer.
We also coalesce memory accesses, fuse scale/bias and cast operations, and
simplify VAE layouts and padding, subject to shape and backend constraints.

\paragraph{Device-specific tuning.}
Quant/GEMM tuning adjusts tile sizes, warps per block, pipeline stages, and buffers.
Backend selection includes VAE layout and convolution tuning
\citep{chetlur2014cudnnefficientprimitivesdeep} and RTX~5090/Spark attention backends.
RTX~5090 retains base Quant/GEMM implementations with shape-specific dispatch
and execution tuning. Spark reduces quantization time with compact buffers
and per-shape choices of warps per block. More resident thread blocks per streaming
multiprocessor (SM) allow the illustrated grid to run in one wave.
Thor's larger per-block shared-memory budget supports GEMM configurations
unavailable on Spark; smaller epilogue tiles avoid register spills.
These adaptations reflect hardware constraints, not device-exclusive algorithms.
Together, shared optimizations and device-specific tuning yield
$3.2$--$4.1\times$ total speedups over BF16 eager execution.
Section~\ref{sec:efficiency} compares model latencies and cumulative gains
(Tables~\ref{tab:efficiency-model-latency} and~\ref{tab:efficiency-cumulative-ablation}).

\begin{figure*}[!t]
\centerline{\includegraphics[width=1.0\textwidth]{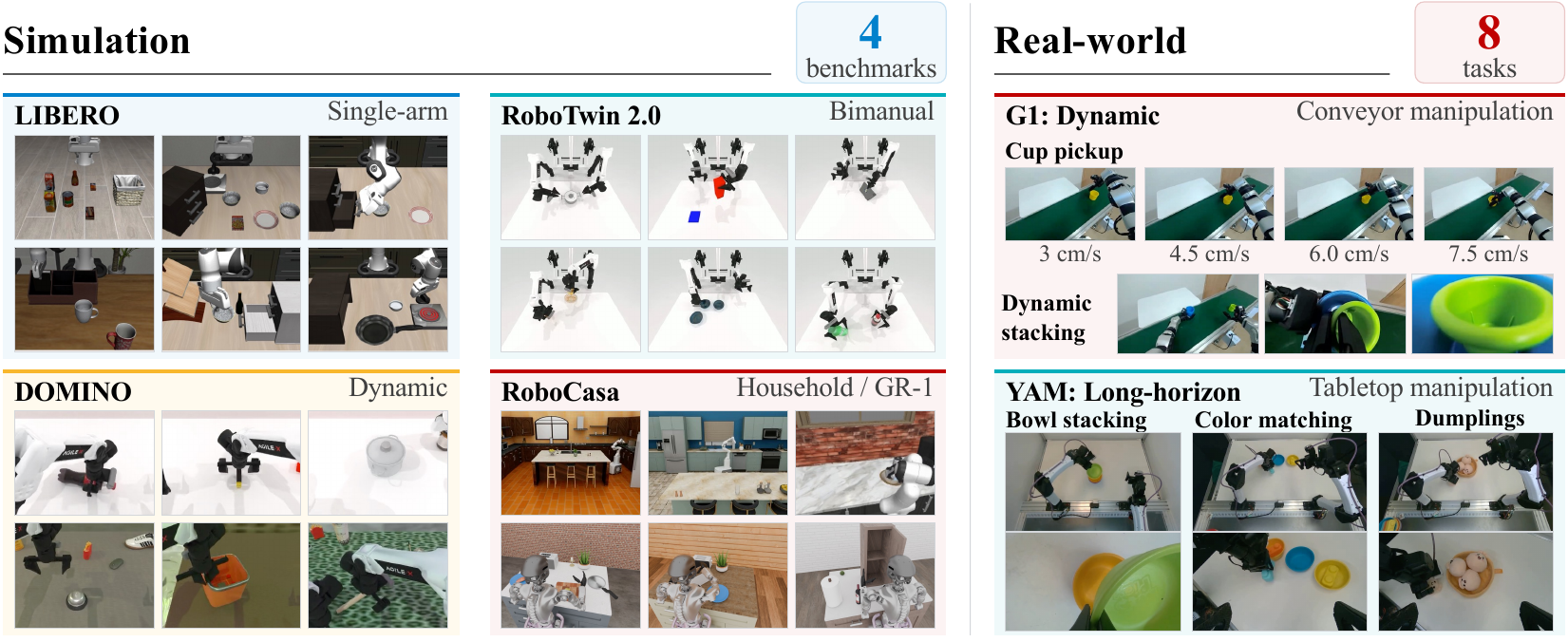}}

\caption{\textbf{Simulation environments and real-world tasks.} Our evaluation spans four simulation benchmarks—LIBERO, RoboTwin 2.0, DOMINO, and RoboCasa—and eight real-world task configurations on G1 and YAM. These include cup pickup at four conveyor speeds, dynamic cup stacking, bowl stacking, brick sorting by color, and dumpling placement into a pan.}
\label{fig:evaluation_overview}
\end{figure*}

\begin{table}[!t]
\centering
\begin{minipage}[t]{0.54\linewidth}
\caption[Success rates on LIBERO]{\raggedright
Success rate (SR, \%) on \textbf{LIBERO}. w/o V: no future-video denoising; CoD: video--action co-denoising; IDM: inverse dynamics modeling.\par}
\label{tab:libero}
\end{minipage}\hfill
\begin{minipage}[t]{0.44\linewidth}
\caption[Success rates on RoboTwin 2.0]{
Success rate (SR, \%) on \textbf{RoboTwin~2.0}. }
\label{tab:robotwin}
\end{minipage}
\par\vspace{3pt}
\begin{minipage}[t]{0.54\linewidth}
\vspace{0pt}
\small
\setlength{\tabcolsep}{2pt}
\renewcommand{\arraystretch}{1.12}
\begin{tabular*}{\linewidth}{@{\extracolsep{\fill}}lccccc@{}}
\toprule
Method & Spatial & Object & Goal & Long & Avg. \\
\midrule
OpenVLA & 84.7 & 88.4 & 79.2 & 53.7 & 76.5 \\
OpenVLA-OFT & 97.6 & 98.4 & 97.9 & 94.5 & 97.1 \\
GR00T-N1 & 94.4 & 97.6 & 93.0 & 90.6 & 93.9 \\
$\pi_0$ & 96.8 & 98.8 & 95.8 & 85.2 & 94.1 \\
$\pi_{0.5}$ & 98.8 & 98.2 & 98.0 & 92.4 & 96.9 \\
UniVLA & 95.4 & 98.8 & 93.6 & 94.0 & 95.5 \\
X-VLA & 98.2 & 98.6 & 97.8 & 97.6 & 98.1 \\
LingBot-VA & 98.5 & 99.6 & 97.2 & 98.5 & 98.5 \\
Motus & 96.8 & 99.8 & 96.6 & 97.6 & 97.7 \\
Fast-WAM & 98.2 & 100.0 & 97.0 & 95.2 & 97.6 \\
\midrule
\textbf{Long-WAM (w/o V)} & 98.0 & 99.5 & 97.0 & 94.5 & 97.3 \\
\textbf{Long-WAM (CoD)} & 98.6 & 99.8 & 96.8 & 95.8 & 97.8 \\
\textbf{Long-WAM (IDM)} & \textbf{99.5} & \textbf{100.0} & \textbf{98.0} & \textbf{99.5} & \textbf{99.5} \\
\bottomrule
\end{tabular*}
\end{minipage}\hfill
\begin{minipage}[t]{0.44\linewidth}
\vspace{0pt}
\small
\setlength{\tabcolsep}{2pt}
\renewcommand{\arraystretch}{1.12}
\begin{tabular*}{\linewidth}{@{\extracolsep{\fill}}lccc@{}}
\toprule
Method & Clean & Rand. & Avg. \\
\midrule
$\pi_0$ & 65.9 & 58.4 & 62.2 \\
$\pi_{0.5}$ & 82.7 & 76.8 & 79.8 \\
Motus & 88.7 & 87.0 & 87.8 \\
Motus (Wan2.2) & 77.6 & 77.0 & 77.3 \\
LingBot-VA & 92.9 & 91.5 & 92.2 \\
LingBot-VA 2.0 & 93.8 & 93.4 & 93.6 \\
Fast-WAM & 91.9 & 91.8 & 91.8 \\
AHA-WAM & 93.4 & 92.2 & 92.8 \\
ABot-M0.5 & 94.0 & 94.2 & 94.1 \\
Qwen-RobotManip & 93.7 & 94.0 & 93.9 \\
\midrule
\textbf{Long-WAM (w/o V)} & 92.4 & 91.6 & 92.0 \\
\textbf{Long-WAM (CoD)} & 94.0 & 93.1 & 93.6 \\
\textbf{Long-WAM (IDM)} & \textbf{94.7} & \textbf{94.2} & \textbf{94.4} \\
\bottomrule
\end{tabular*}
\end{minipage}
\end{table}

\section{Experiments}
\label{sec:experiments}

\subsection{Implementation Details}
\label{sec:implementation}

Pretraining LongLive2.0-Robot uses approximately 30,720 GPU-hours on
64 NVIDIA H100 GPUs; downstream WAMs train on 16 GB200 GPUs.
WAM training uses AdamW with a peak learning rate of $10^{-4}$, cosine decay,
5\% warmup, BF16 precision, and gradient clipping at 1.0.
Robot experiments use a Unitree G1 humanoid and a YAM bimanual manipulator.
Appendix~\ref{app:baseline-details} lists baseline references.

\subsection{Results on Simulation Benchmarks}
\label{sec:simulation}

We evaluate on LIBERO \citep{liu2023libero}, RoboTwin~2.0
\citep{chen2025robotwin2}, DOMINO \citep{fang2026domino}, and RoboCasa GR-1
\citep{nvidiagear2025gr1tabletop} (Figure~\ref{fig:evaluation_overview}). For the
first three benchmarks, the main results use up to 2.4 seconds of context;
DOMINO adds moving objects, where recent motion informs anticipation
and interception. RoboCasa GR-1, whose tasks chain several object transfers,
hosts the longer-context study. We further evaluate compositional tasks on RoboCasa365, with
GPT-6 Astra as a high-level planner
(Appendix~\ref{app:robocasa365-planning}, Table~\ref{tab:robocasa365-planning}).


\paragraph{LIBERO.}
\label{sec:sim-libero}
Long-WAM (IDM) achieves the highest average success (99.5\%) and
99.5\% on LIBERO-Long (Table~\ref{tab:libero}). Its margin is largest on LIBERO-Long, the suite with the longest tasks, where it exceeds LingBot-VA \citep{li2026lingbotva} and Fast-WAM \citep{yuan2026fastwam} by 1.0 and 4.3 points.

\begin{wraptable}{r}{0.4\textwidth}
\vspace{-35pt}
\centering
\caption{DOMINO results after dynamic-data fine-tuning. Success rate (SR);
manipulation score (MS).}
\label{tab:domino-ft}
\small
\setlength{\tabcolsep}{4pt}
\renewcommand{\arraystretch}{1.12}
\begin{tabular*}{\linewidth}{@{\extracolsep{\fill}}lcc@{}}
\toprule
Method & SR (\%) $\uparrow$ & MS $\uparrow$ \\
\midrule
OpenVLA & 1.5 & 6.1 \\
$\pi_0$ & 8.2 & 24.0 \\
$\pi_{0.5}$ & 9.6 & 26.2 \\
InternVLA-M1 & 5.4 & 27.6 \\
OpenVLA-OFT & 9.1 & 24.1 \\
StarVLA-OFT & 10.9 & 30.5 \\
PUMA & 17.2 & 35.0 \\
Fast-WAM & 19.9 & 33.3 \\
\midrule
\textbf{Long-WAM} & \textbf{34.9} & \textbf{45.1} \\
\bottomrule
\end{tabular*}
\vspace{-45pt}
\end{wraptable}

\paragraph{RoboTwin 2.0.}
\label{sec:sim-robotwin}
Long-WAM (IDM) leads on average (94.4\%) and Clean (94.7\%), and ties the
best Randomized score (94.2\%; Table~\ref{tab:robotwin}). Its average
exceeds ABot-M0.5 and LingBot-VA~2.0 \citep{zhang2026lingbotva2} by
0.3 and 0.8 points, respectively.

\paragraph{DOMINO.}
\label{sec:sim-domino}

After dynamic-data fine-tuning, Long-WAM leads with 34.9\% SR and 45.1 MS
(Table~\ref{tab:domino-ft}), exceeding the strongest baseline on each metric:
Fast-WAM by 15.0 SR points and PUMA \citep{fang2026domino} by 10.1 MS points.
Recent observations supply motion cues that a single image cannot, supporting prediction-conditioned interception; Section~\ref{sec:real-world} tests this ability on a physical conveyor.

\subsection{Ablation Studies}
\label{sec:ablations}
\label{sec:scaling-experiments}

\begin{figure}[t]
\centering
\begin{minipage}[t]{0.72\textwidth}
\vspace{0pt}
\caption{Context scaling with three video initializations. Context windows
are displayed at equal spacing.}
\label{fig:memory-scaling}
\end{minipage}\hfill
\begin{minipage}[t]{0.27\textwidth}
\vspace{0pt}
\makeatletter
\def\@captype{table}
\makeatother
\caption{Success rate (SR, \%) on RoboCasa GR-1.}
\label{tab:gr1-comparison}
\end{minipage}
\par\vspace{3pt}
\begin{minipage}[t]{0.72\textwidth}
\vspace{0pt}
\centering
\includegraphics[width=\linewidth,trim=0bp 7bp 0bp 5.5bp,clip]{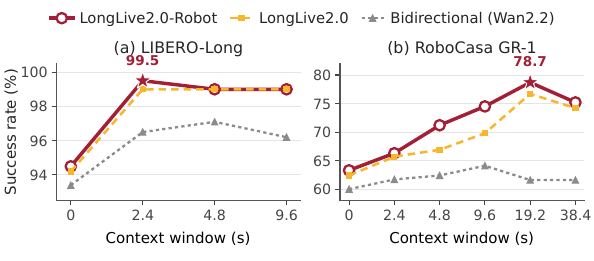}
\end{minipage}\hfill
\begin{minipage}[t]{0.27\textwidth}
\vspace{0pt}
\centering
\small
\setlength{\tabcolsep}{2pt}
\renewcommand{\arraystretch}{1.00}
\begin{tabular*}{\linewidth}{@{\extracolsep{\fill}}lr@{}}
\toprule
Method & SR $\uparrow$ \\
\midrule
$\pi_0$ & 62.5 \\
Diffusion Policy & 32.7 \\
Fast-WAM & 47.5 \\
GR00T-N1.5 & 64.1 \\
DreamZero & 62.4 \\
Cosmos Policy & 67.1 \\
\midrule
\textbf{Long-WAM (2.4 s)} & \textbf{63.3} \\
\textbf{Long-WAM (19.2 s)} & \textbf{78.7} \\
\bottomrule
\end{tabular*}
\end{minipage}
\end{figure}

\begin{table}[t]
\centering
\caption{\textbf{Success rate (SR, \%) on RoboCasa365.} Overall averages 50 tasks:
18 Atomic-Seen, 16 Composite-Seen, and 16 Composite-Unseen. Baseline sources: Appendix~\ref{app:baseline-details}.}
\label{tab:robocasa365-planning}
\small
\setlength{\tabcolsep}{4pt}
\renewcommand{\arraystretch}{1.08}
\begin{tabular*}{\linewidth}{@{\extracolsep{\fill}}lcccc@{}}
\toprule
Method & \shortstack{Atomic\\Seen} & \shortstack{Composite\\Seen}
& \shortstack{Composite\\Unseen} & Overall \\
\midrule
Diffusion Policy & 15.7 & 0.2 & 1.3 & 6.1 \\
Azero-Robotics-1 & 30.3 & 3.8 & 1.6 & 12.6 \\
$\pi_0$ & 34.6 & 6.1 & 1.1 & 14.8 \\
GigaWorld-Policy 0.1 & 44.4 & 11.8 & 2.9 & 20.7 \\
GR00T N1.6 & 51.1 & 9.4 & 1.7 & 21.9 \\
GR00T N1.5 & 50.7 & 14.8 & 2.7 & 23.9 \\
WorldDreamer & 66.3 & 26.7 & 9.0 & 35.3 \\
RLDX-1 & 67.6 & 27.9 & 8.5 & 36.0 \\
PRTS & 66.3 & 30.3 & 18.8 & 39.6 \\
Qwen-RobotManip	&68.6	&20.1	&14.9	&35.9 \\
ABot-M0.5 & 75.6 & 37.7 & 3.3 & 40.3 \\
\midrule
GPT-6 Astra & 31.5 & 22.4 & 20.8 & 25.2 \\
\hdashline
$\pi_{0.5}$ & 39.6 & 7.1 & 1.2 & 16.9 \\
$\pi_{0.5}$ + GPT-6 Astra & 46.5 & 24.0 & 19.0 & 30.5 \\
\hdashline
\textbf{Long-WAM} & 67.9 & 15.8 & 6.1 & 31.4 \\
\textbf{Long-WAM + GPT-6 Astra} & \textbf{85.6} & \textbf{38.8}
& \textbf{35.0} & \textbf{54.4} \\
\bottomrule
\end{tabular*}
\par\smallskip
\end{table}

\paragraph{Context Scaling.}

Adding 2.4 seconds of history raises LIBERO-Long success from 94.5\% to
99.5\% with LongLive2.0-Robot and 94.2\% to 99.0\% with LongLive-2.0
(Figure~\ref{fig:memory-scaling}). On RoboCasa GR-1, chosen for its
multi-stage tasks \citep{nasiriany2024robocasa,nvidiagear2025gr1tabletop},
scaling from 2.4 to 19.2 seconds improves success from 66.3\% to 78.7\%
and 65.7\% to 76.7\%, respectively. The different peak context lengths suggest
task-dependent memory needs: short histories capture most gains on LIBERO-Long,
while GR-1 benefits from substantially longer interaction context.
Robot-video pretraining yields higher peaks on both benchmarks; its 78.7\% exceeds the strongest GR-1 baseline
by 11.6 points (Table~\ref{tab:gr1-comparison}). At 38.4
seconds, success decreases to 75.2\% and 74.2\%; this window is three times the
average training trajectory (12.1 seconds), and 80.4\% of its sampled history
frames are padding. We hypothesize that the decline reflects limited history
coverage rather than an intrinsic memory limit.
Context also has a price: 8$\times$ more history raises
RTX~5090 latency 3.2$\times$ (107.4 to 341.0\,ms; Appendix~\ref{app:context-latency}).

\paragraph{Autoregressive vs. Bidirectional Pretraining.}
\label{sec:pretraining-ablation}

The AR advantage itself grows with context: on GR-1, the robot-domain AR
variant leads bidirectional initialization by 3.3 points without history and
17.1 points at 19.2 seconds (Figure~\ref{fig:memory-scaling}). The
bidirectional variant rises from 61.7\% at 2.4 seconds to 64.1\% at 9.6,
then returns to 61.6\% at 19.2; both AR variants instead gain 12.4 and
11.0 points over the 2.4--19.2-second interval.
All variants use causal WAM adaptation, yet access to history alone does
not reproduce the AR variants' long-context gains. A plausible explanation is
that AR pretraining learns the history-to-future dependencies retained
during action adaptation, allowing additional observations to inform control
through a predictive representation.

\begin{figure}[t]
\centering
\includegraphics[width=\textwidth]{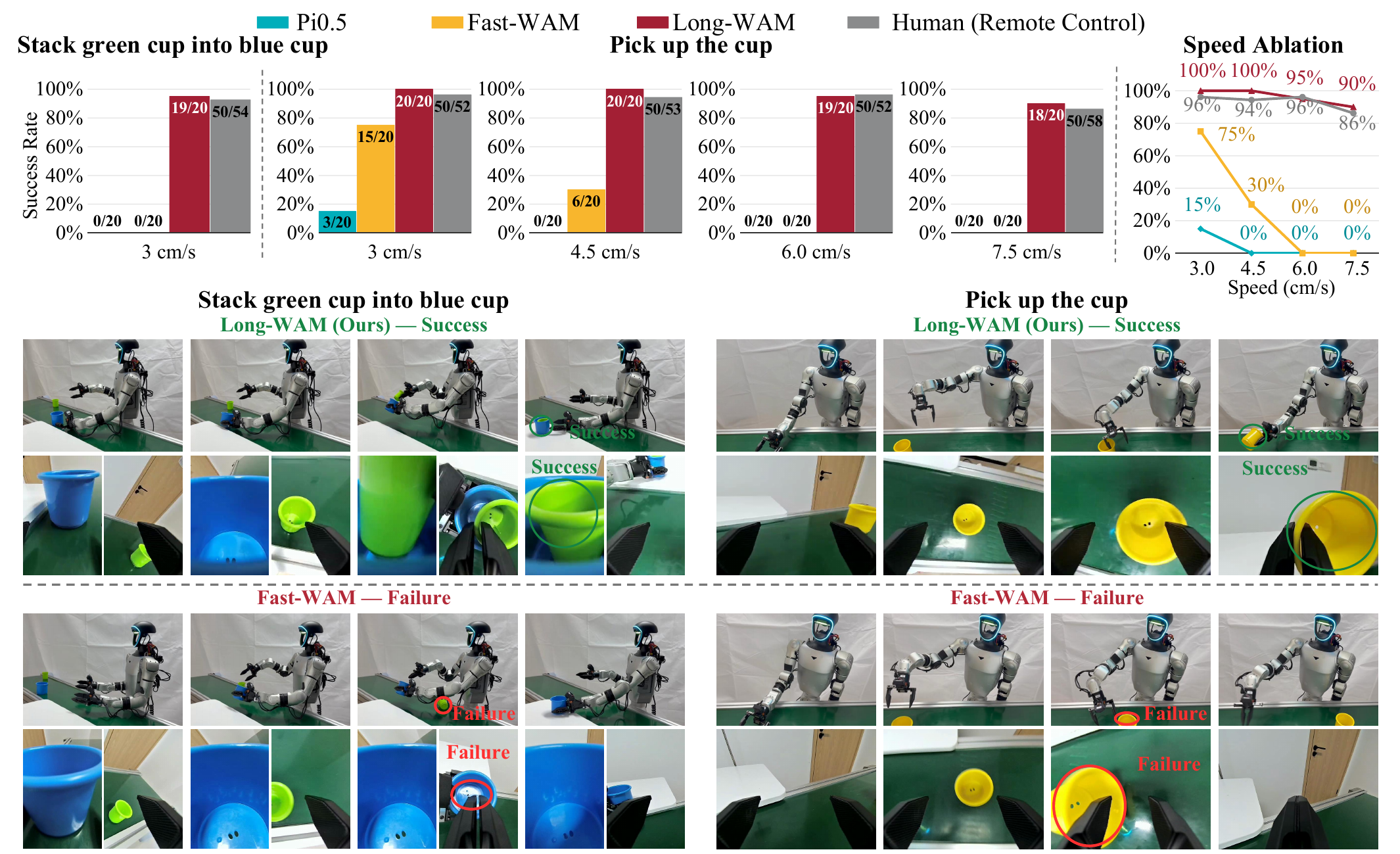}
\caption{\textbf{Dynamic manipulation on Unitree G1.} Long-WAM maintains
90--100\% grasping success across conveyor speeds and achieves 95\% success
on dynamic cup stacking. Top: policy and human teleoperation comparisons. Bottom: Long-WAM and Fast-WAM rollouts.}
\label{fig:real-dynamic}
\end{figure}

\paragraph{Video--Action Denoising Strategy.}
\label{sec:denoising-ablation}
We compare action denoising without future-video prediction (\emph{w/o V}),
joint video--action denoising (\emph{CoD}), and video prediction followed by
history- and prediction-conditioned action denoising (\emph{IDM};
Tables~\ref{tab:libero} and~\ref{tab:robotwin}). IDM's largest suite-level
gains occur on LIBERO-Long: 99.5\%, versus 94.5\% (w/o V) and 97.8\% (CoD),
suggesting that first estimating how an interaction will evolve provides a
useful condition for coordinating subsequent actions across multiple substeps.
The resulting policy also supports dynamic grasping
(Section~\ref{sec:real-world}). These gains use partially denoised future
latents, without pixel-level synthesis, highlighting prediction as an
intermediate control representation. Our infrastructure addresses the
sequential-inference cost while preserving this predictive path
(Section~\ref{sec:efficiency}).

\section{Atomic Execution and Compositional Planning}
\label{app:robocasa365-planning}

RoboCasa365 \citep{nasiriany2026robocasa365} separates atomic household skills
from their composition into multi-stage tasks, allowing us to examine
Long-WAM as the execution foundation of a hierarchical robot system.
We use a Human300-trained checkpoint with 2.4 seconds of visual context,
then add GPT-6 Astra without further policy training. The planner grounds
task goals into atomic subinstructions, selects execution-prefix lengths,
and can issue bounded end-effector corrections; Long-WAM supplies the
learned action chunks. Table~\ref{tab:robocasa365-planning} compares GPT-6 Astra
alone, standalone and planner-augmented policies, and benchmark reference
methods \citep{robocasateam2026leaderboard}.

\paragraph{Strong atomic execution.}
Long-WAM alone achieves 67.9\% Atomic-Seen success, compared with 39.6\%
for $\pi_{0.5}$ and 31.5\% for GPT-6 Astra alone, providing a strong
physical skill foundation. Querying the same policy every 15 steps yields
84.4\% without a planner, close to the hierarchical system's 85.6\%.

\paragraph{Planning unlocks unseen skill compositions.}
With the policy checkpoint and visual context unchanged, adding the planner
raises Composite-Seen success from 15.8\% to 38.8\% and Composite-Unseen
success from 6.1\% to 35.0\%; Overall improves from 31.4\% to 54.4\%.
The 15-step control reaches only 11.2\% and 5.0\% on the composite splits,
so shorter execution intervals alone do not explain these gains.

\paragraph{Strong policies amplify agentic planning.}
GPT-6 Astra alone reaches 25.2\% Overall success, and pairing it with
$\pi_{0.5}$ reaches 30.5\%, compared with 54.4\% for Long-WAM + GPT-6 Astra.
On unseen compositions, the Long-WAM hierarchy achieves 35.0\%, exceeding
both GPT-6 Astra alone (20.8\%) and the $\pi_{0.5}$ hierarchy (19.0\%).
Planner augmentation yields a larger reported Overall gain for Long-WAM:
23.0 percentage points, versus 13.6 for $\pi_{0.5}$.
These system-level comparisons highlight execution
quality as a key complement to reasoning: Long-WAM supplies strong physical
skills, while high-level planning extends their use to unseen compositions.
This supports the division of labor in Appendix~\ref{sec:discussion}, in which
agentic planning and memory-informed execution jointly enable more capable
long-horizon behavior.

\subsection{Real-world Deployment}
\label{sec:real-world}
\begin{wrapfigure}{r}{0.35\textwidth}
\vspace{-45pt}
\centering
\includegraphics[width=\linewidth]{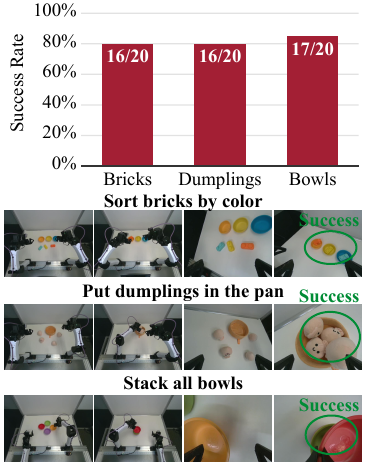}
\vspace{-20pt}
\caption{\textbf{Long-horizon execution on YAM:} 20 trials per task.}
\label{fig:real-long-horizon}
\vspace{-60pt}
\end{wrapfigure}

We evaluate dynamic manipulation on Unitree G1 and long-horizon tasks on YAM,
with 20 trials per policy and condition.

\paragraph{Dynamic Tasks.}
\label{sec:real-dynamic-speed}
Conveyor speed separates the policies (Figure~\ref{fig:real-dynamic}). As
the belt accelerates from 3.0 to 7.5~cm/s, the grasping success of Fast-WAM
\citep{yuan2026fastwam} falls from 75\% to 0\% and that of $\pi_{0.5}$
\citep{physicalintelligence2025pi05} from 15\% to 0\%, whereas Long-WAM stays
at 90--100\% (100\%, 100\%, 95\%, and 90\%).
\label{sec:real-dynamic-composite}Stacking a moving green cup into a blue cup
at 3~cm/s adds alignment and placement to interception; Long-WAM succeeds in
19 of 20 trials, and neither baseline succeeds once.

\paragraph{Long-Horizon Tasks.}
\label{sec:real-long-horizon}

On YAM, where tasks last over 40 seconds on average, Long-WAM succeeds in
80\%, 80\%, and 85\% of trials on brick sorting by color, placing dumplings in
a pan, and stacking bowls (81.7\% on average; Figure~\ref{fig:real-long-horizon}),
sustaining multi-step execution and rapid interception.


\subsection{Efficiency}
\label{sec:efficiency}


\begin{table}[t]
\centering
\small
\setlength{\tabcolsep}{2pt}
\setlength{\abovecaptionskip}{3pt}
\setlength{\belowcaptionskip}{3pt}
\newsavebox{\efficiencyLatencyPanel}
\begin{lrbox}{\efficiencyLatencyPanel}
\begin{minipage}[t]{0.49\textwidth}
\vspace{0pt}
\centering
\caption{RTX~5090 end-to-end latency and RoboTwin~2.0 success rate (SR; Clean/Randomized mean).
BF16 eager is the unoptimized V4/A4 baseline.}
\label{tab:efficiency-model-latency}
\footnotesize
\begin{tabular*}{\linewidth}{@{\extracolsep{\fill}}lrr@{}}
\toprule
Method & Latency (ms) $\downarrow$ & SR (\%) $\uparrow$ \\
\midrule
LingBot-VA & 3618.4 & 92.2 \\
Motus & 1201.1 & 87.8 \\
Cosmos Policy & 470.6 & -- \\
Fast-WAM & 244.1 & 91.8 \\
\midrule
\multicolumn{3}{@{}l}{\textbf{Long-WAM}} \\
BF16 eager & 356.0 & 94.4 \\
\textbf{Optimized (V4/A4)} & \textbf{107.4} & \textbf{93.5} \\
\textbf{Optimized (V2/A2)} & \textbf{81.8} & \textbf{92.5} \\
\bottomrule
\end{tabular*}
\par\vspace{0pt}
\end{minipage}
\end{lrbox}
\usebox{\efficiencyLatencyPanel}\hfill
\begin{minipage}[t][\dimexpr\ht\efficiencyLatencyPanel+\dp\efficiencyLatencyPanel\relax][s]{0.49\textwidth}
\vspace{0pt}
\centering
\caption{RoboTwin~2.0 success rate (SR, \%). Sync: Table~\ref{tab:robotwin}; Async: our evaluation.
Jerk denotes a discrete smoothness proxy; see Appendix~\ref{app:async-overlap}.}
\label{tab:async-continuity}
\par\vfill
\footnotesize
\renewcommand{\arraystretch}{1.50333}
\begin{tabular*}{\linewidth}{@{\extracolsep{\fill}}lrrrr@{}}
\toprule
Method & Sync & \multicolumn{3}{c}{Async} \\
\cmidrule(lr){2-2}\cmidrule(l){3-5}
& SR $\uparrow$ & SR $\uparrow$ & RMSE $\downarrow$ & Jerk $\downarrow$ \\
\midrule
Fast-WAM & 91.8 & 76.4 & 0.0731 & 0.1316 \\
LingBot-VA & 92.2 & 46.7 & 0.1432 & 0.6592 \\
Long-WAM (CoD) & 93.6 & 93.2 & 0.0260 & 0.0479 \\
\textbf{Long-WAM (IDM)} & \textbf{94.4} & \textbf{94.2} & \textbf{0.0246} & \textbf{0.0436} \\
\bottomrule
\end{tabular*}
\par\vspace{0pt}
\end{minipage}
\end{table}

\paragraph{Latency.}
On RTX~5090, the optimizations detailed in Section~\ref{sec:edge-acceleration} jointly reduce
the V4/A4 end-to-end inference latency to 107.4\,ms while largely preserving SR
(BF16: 94.4\%; optimized: 93.5\%; Table~\ref{tab:efficiency-model-latency}). Long-WAM achieves a
$2.3\times$ speedup over Fast-WAM \citep{yuan2026fastwam} (244.1\,ms) despite additionally predicting future video frames.
Both optimized variants outperform existing baselines in both inference latency and reported SR;
notably, V2/A2 trades a modest 1.0-point drop in SR for an additional 24\% latency reduction (81.8\,ms).
The other baselines are evaluated under their native deployment configurations and denoising budgets.

\par\addvspace{10pt}
\begingroup
\small
\setlength{\tabcolsep}{2pt}
\setlength{\abovecaptionskip}{3pt}
\setlength{\belowcaptionskip}{3pt}
\noindent
\begin{minipage}{0.98\textwidth}
\centering
\makeatletter
\def\@captype{table}
\makeatother
\caption{End-to-end latency (E2E) of Long-WAM under cumulative optimizations across devices, including the full VAE computation. Each row keeps all preceding optimizations on a fixed input; speedups are relative to BF16 eager. The first optimized stage combines NVFP4 quantization, CUDA Graph replay, and compilation.}
\label{tab:efficiency-cumulative-ablation}
\begin{tabular*}{\linewidth}{@{\extracolsep{\fill}}lrrrrrr@{}}
\toprule
Optimization & \multicolumn{2}{c}{RTX 5090} & \multicolumn{2}{c}{DGX Spark} & \multicolumn{2}{c}{Thor} \\
\cmidrule(lr){2-3}\cmidrule(lr){4-5}\cmidrule(l){6-7}
& E2E (ms) $\downarrow$ & Speedup $\uparrow$ & E2E (ms) $\downarrow$ & Speedup $\uparrow$ & E2E (ms) $\downarrow$ & Speedup $\uparrow$ \\
\midrule
\multicolumn{7}{@{}l}{\textbf{Shared optimizations}} \\
BF16 eager & 356.0 & 1.0$\times$ & 1342.8 & 1.0$\times$ & 1215.2 & 1.0$\times$ \\
\shortstack[l]{+ NVFP4 quantization\\\quad + CUDA Graph + compile} & 172.4 & 2.1$\times$ & 686.4 & 2.0$\times$ & 762.4 & 1.6$\times$ \\
+ Denoising-invariant reuse & 144.9 & 2.5$\times$ & 464.5 & 2.9$\times$ & 524.5 & 2.3$\times$ \\
+ Shared input quantization & 130.3 & 2.7$\times$ & 427.8 & 3.1$\times$ & 476.5 & 2.6$\times$ \\
+ Video--action online softmax & \textbf{126.9} & \textbf{2.8$\times$} & \textbf{419.9} & \textbf{3.2$\times$} & \textbf{468.2} & \textbf{2.6$\times$} \\
\midrule
\multicolumn{7}{@{}l}{\textbf{Device-specific tuning}} \\
+ Quant/GEMM tuning & 119.1 & 3.0$\times$ & 415.6 & 3.2$\times$ & 458.9 & 2.6$\times$ \\
+ Backend selection & \textbf{107.4} & \textbf{3.3$\times$} & \textbf{328.2} & \textbf{4.1$\times$} & \textbf{378.7} & \textbf{3.2$\times$} \\
\bottomrule
\end{tabular*}
\end{minipage}
\par
\endgroup

\paragraph{Pure Asynchronous Execution.}
Long-WAM largely preserves synchronous success. At $S=12$ ($R=24$), IDM
loses 0.2 percentage points (94.4\% to 94.2\%) and CoD loses 0.4;
Fast-WAM and LingBot-VA \citep{li2026lingbotva} lose 15.4
and 45.5 (Table~\ref{tab:async-continuity}). Long-WAM's overlap RMSE and jerk are about one-third of Fast-WAM's. With a shorter overlap ($S=16$), asynchronous success reaches
95.0\%, on par with synchronous execution (Appendix~\ref{app:async-overlap}).

\paragraph{Cumulative Acceleration.}
NVFP4's compute savings can be offset by the launch overhead of added small kernels.
When combined with CUDA Graph replay and compilation, NVFP4 yields end-to-end
speedups while retaining reduced weight and activation storage.
Together with the remaining shared optimizations, this reduces latency to
126.9, 419.9, and 468.2\,ms on RTX~5090, Spark, and Thor
(Table~\ref{tab:efficiency-cumulative-ablation}).
Device-specific tuning reduces latency by another 15--22\%, yielding
107.4, 328.2, and 378.7\,ms, for total speedups of
$3.3\times$, $4.1\times$, and $3.2\times$ over BF16 eager.
Appendix~\ref{app:idm-codenoise} details timing and configuration-dependent KV reuse effects.


\clearpage
{
  \small
  \bibliography{iclr2027_conference}
  \bibliographystyle{iclr2027_conference}
}

\clearpage
\appendix

\section{Discussion and Conclusion}
\label{sec:discussion}

Long-WAM studies context scaling where memory must ultimately serve action.
The empirical gains on LIBERO-Long, RoboCasa GR-1, and real-world manipulation
make recent physical history a useful resource for world-action modeling,
while the deployment system connects that resource to responsive execution.
This shifts the design question from whether a policy has memory to how its
context, predictive representation, and execution budget should be designed
together.

The goal is not unlimited history in the low-level controller.  A manipulation
policy benefits from remembering motion and recent interaction state, whereas
hour-scale goals, reasoning, and task decomposition are naturally handled by
a higher-level planner.  These roles are complementary: a planner can maintain
semantic continuity while Long-WAM executes local objectives with the physical
context needed for control.  The appropriate window may therefore depend on
the task and the available compute, rather than follow a universal duration.

Each context window in our study is trained as its own model, which gives a
clean view of how much history helps at every length; the natural next step
is a single policy that adjusts its window at test time. The measured latency
profile (74.6\,ms without history, 341.0\,ms at 19.2 seconds;
Appendix~\ref{app:context-latency}) indicates how much computation such
adaptivity could reclaim. The decline at 38.4 seconds coincides with sparse real
history in current datasets (80.4\% padding), so longer and denser robot
recordings offer a direct route to extending the useful window. With the
streaming deployment stack in place, real-robot context-length studies are a
natural next step, showing how the simulated trends carry over to physical
interaction.

The observed trends motivate adaptive context allocation: preserving the
history relevant to the current interaction while matching computation to its
response requirements.  They do not prescribe a power law or a fixed memory
optimum.  More broadly, Long-WAM motivates evaluating generative robot policies
jointly by what they remember, how well they act, and how quickly they can
incorporate new observations.

\section{Data and Model Training Details}

\subsection{Pretraining Data}
\label{app:data}

The pretraining corpus contains 2,294,889 model-ready text-and-image-to-video
samples from five dataset families.  Table~\ref{tab:pretraining-data} records
the six source subsets, including two AgiBot World subsets.  The reported
training, validation, and test splits contain 2,251,021, 21,931, and 21,937
samples, respectively.  The approximately 10,000-hour training scale is a
window-equivalent estimate: assigning 16 seconds to each training sample
yields 10,004.5 aggregate hours.  It is not a measurement of unique raw footage;
overlapping windows, source clip lengths, and padding affect that distinction.

\begin{table}[ht]
\centering
\caption{Pretraining data composition. Counts describe the full corpus before
the train/validation/test split, not the published size of each source dataset.}
\label{tab:pretraining-data}
\begin{tabular}{lr}
\toprule
Source subset & Samples \\
\midrule
RoVid-X \citep{deng2026rovidx} & 1,898,562 \\
AgiBot World: \texttt{full146} \citep{agibotworld2025} & 69,834 \\
AgiBot World: 5.2--32 s supplement & 40,497 \\
EgoDex \citep{hoque2026egodex} & 207,044 \\
EgoVerse \citep{punamiya2026egoverse} & 59,742 \\
VITRA \citep{li2025vitra} & 19,210 \\
\midrule
Total & 2,294,889 \\
\bottomrule
\end{tabular}
\end{table}

Pretraining uses video prediction without action supervision, so sources need
not share action coordinates or actuator dimensions.  Action-space-independent
supervision is a property of this stage; improved transfer to an unseen
embodiment would require a separate evaluation.

\subsection{Training Objectives and Inference Interface}
\label{app:training}

\paragraph{Video pretraining.}
Let $z_i$ denote an uncorrupted target chunk and $\epsilon_i\sim\mathcal N(0,I)$
its base Gaussian noise.  With context, latent, and noise perturbations
$\delta_i^c$, $\delta_i^z$, and $\delta_i^\epsilon$, the error-recycling inputs are
\begin{equation}
\bar\epsilon_i=\epsilon_i+\delta_i^\epsilon,\qquad
x_i^{\sigma_i}=(1-\sigma_i)(z_i+\delta_i^z)+\sigma_i\bar\epsilon_i,
\qquad h_{<i}=(z_j+\delta_j^c)_{j<i}.
\end{equation}
Perturbations are sampled from buffered model errors when their corresponding
augmentation is enabled, and are zero otherwise.  The velocity target is
$\bar\epsilon_i-z_i$, not $\bar\epsilon_i-(z_i+\delta_i^z)$: latent-input
corruption is corrected toward the original target.  The conditioning image
remains unperturbed and is excluded from the loss.  Paired clean/noisy streams
implement teacher forcing: a target block accesses preceding context blocks
and its own noisy tokens, without accessing later blocks or its own clean
target.  Loss is averaged over eligible target elements and weighted by the
scheduler, as in Eq.~\ref{eq:ar-video}.

\paragraph{World-action adaptation.}
Both branches use the flow path $x^\sigma=(1-\sigma)x+\sigma\epsilon$ and target
$u=\epsilon-x$.  For branch $b\in\{v,a\}$, let $m_b$ select valid supervised
elements and $r_b$ be its prediction residual.  The masked loss is
\begin{equation}
\mathcal L_b=\mathbb E\!\left[
w_b(\sigma_b)\frac{\|m_b\odot r_b\|_2^2}{\max(\|m_b\|_1,1)}\right],
\qquad r_b=v_b(x_b^{\sigma_b},\sigma_b\mid\text{conditioning})-(\epsilon_b-x_b).
\end{equation}
Here $\odot$ is element-wise multiplication, and the temporal validity mask
is broadcast over feature dimensions.  It excludes fully padded future
latent steps and padded action timesteps.  The video pass supervises noisy
future latents with the observed prefix clamped clean.  The action pass uses
a detached video cache built from the observed prefix and ground-truth future
latents forward-noised to $\sigma_\star=0.9$.  It updates the action expert and
the shared proprioceptive adapter through the action-conditioning path, without
backpropagating into the video cache.  The two-pass objective does not use the
paired-stream or error-recycling augmentations of video pretraining.

At inference, partial video rollout starts from Gaussian noise and stops at
$\sigma_\star$; it does not observe ground-truth future frames.  Training and
inference share this noise level but not the source of the future latents:
the training surrogate retains a residual ground-truth component.  This is a
training approximation, not an assertion of identical conditioning
distributions.  Future latents remain in latent space, and the resulting
visual KV cache is reused throughout joint action-chunk denoising.

\begin{figure*}[t]
\centering
\includegraphics[width=\linewidth]{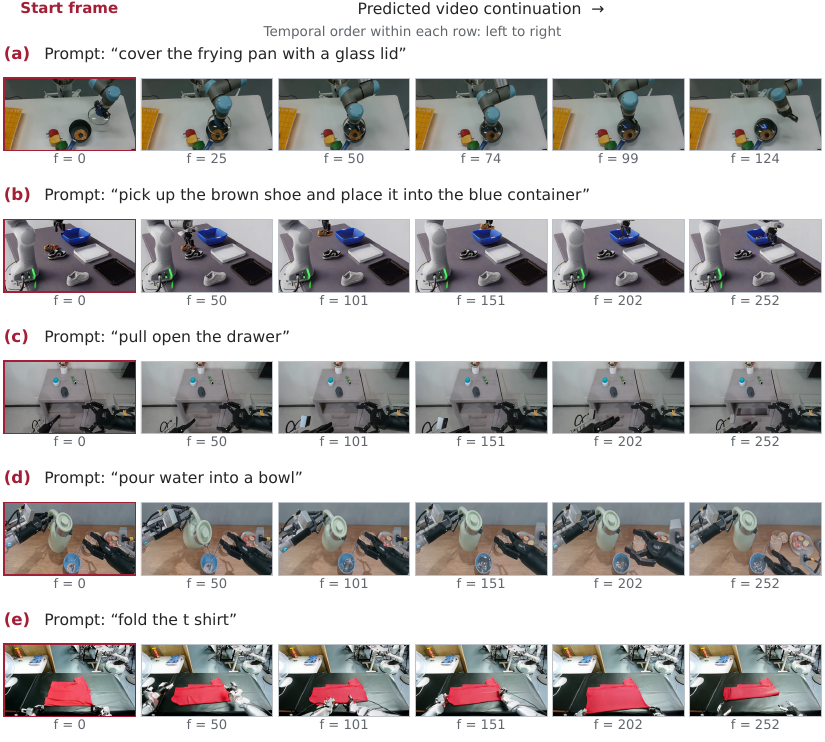}
\caption{\textbf{Task-conditioned video prediction with LongLive2.0-Robot.}
Five selected text-and-image-to-video (TI2V) examples, with the task prompt
shown above each row. $f$ denotes the zero-based frame index. All frames retain the original field of view.}
\label{fig:ti2v-predictions}
\end{figure*}

\paragraph{Cache scope.}
The implementation appends a projection of the latest robot state
$q_t$ to the conditioning tokens of both experts.  Consequently, visual
features can depend on $q_t$ at multiple layers.  Reuse within one action solve
is distinct from reuse across decisions: updating $q_t$, evicting old context,
or changing temporal positions can invalidate previously computed visual
keys and values.  Any persistent-cache optimization must specify its refresh
semantics and maintain the intended observation/position alignment.

\section{LongLive2.0-Robot Video Predictions}
\label{app:video-predictions}


Given a single image and a language instruction, LongLive2.0-Robot predicts
how a robot interacts with its environment (Figure~\ref{fig:ti2v-predictions}).
The selected rollouts follow the prompted tasks: the gripper
positions a lid over a pan and withdraws, transfers
the specified shoe into a container, and pulls a drawer open. Pouring and
T-shirt folding further illustrate coordinated motion involving changing
object configurations. Across these examples, the predicted arm movements
and object transitions form coherent, task-directed sequences rather than
merely preserving scene appearance. These qualitative results provide
evidence that robot-video pretraining learns a predictive representation of
robot motion and object interaction, supplying a task-conditioned visual
prior for subsequent world-action adaptation.

\section{Baseline Details}
\label{app:baseline-details}

Our policy baselines include Diffusion Policy \citep{chi2024diffusionpolicyvisuomotorpolicy},
OpenVLA \citep{kim2024openvlaopensourcevisionlanguageactionmodel},
OpenVLA-OFT \citep{kim2025finetuningvisionlanguageactionmodelsoptimizing},
GR00T \citep{nvidia2025gr00tn1openfoundation},
$\pi_0$ \citep{black2026pi0visionlanguageactionflowmodel},
$\pi_{0.5}$ \citep{physicalintelligence2025pi05},
UniVLA \citep{wang2025unifiedvisionlanguageactionmodel}, X-VLA \citep{zheng2025xvla},
InternVLA-M1 \citep{chen2025internvlam1spatiallyguidedvisionlanguageaction}, and
StarVLA \citep{community2026starvlalegolikecodebasevisionlanguageaction}.
UniVLA refers to Wang et al.'s \emph{Unified Vision-Language-Action Model};
PUMA is the method introduced with DOMINO \citep{fang2026domino}.

World-action baselines include Motus \citep{bi2025motus},
LingBot-VA \citep{li2026lingbotva}, LingBot-VA 2.0 \citep{zhang2026lingbotva2},
Fast-WAM \citep{yuan2026fastwam}, AHA-WAM \citep{cai2026ahawam},
DreamZero \citep{ye2026dreamzero}, Cosmos Policy \citep{kim2026cosmos}, and
ABot-M0.5 \citep{chen2026abotm05unifiedmobilityandmanipulationworld}.
RoboCasa365 comparisons additionally include Azero-Robotics-1, WorldDreamer,
GR00T N1.5/N1.6,
GigaWorld-Policy \citep{ye2026gigaworldpolicyefficientactioncenteredworldaction},
RLDX-1 \citep{kim2026rldx1technicalreport}, and
PRTS \citep{zhang2026prtsprimitivereasoningtasking}, with external baseline scores
from the leaderboard \citep{robocasateam2026leaderboard}, except Qwen-RobotManip,
whose scores come from its technical report \citep{yuan2026qwenrobotmaniptechnicalreportalignment}.

Video initialization comparisons use Wan2.2 \citep{wan2025},
LongLive-2.0 \citep{chen2026longlive2}, and our LongLive2.0-Robot;
GPT-6 Astra serves as the high-level planner in Appendix~\ref{app:robocasa365-planning}.

\section{Discussion of Asynchronous Overlap Length}
\label{app:async-overlap}

We evaluate Long-WAM (IDM) on RoboTwin~2.0 \citep{chen2025robotwin2}, varying
the trigger stride $S$ with the execution horizon fixed at
$R=24$, giving overlaps $O=R-S$ of 12, 8, and 4 control steps.
Table~\ref{tab:async-overlap} reports Long-WAM's success rate and action
continuity under these settings. The $S=12$ result is also used for
Long-WAM (IDM) in Table~\ref{tab:async-continuity}.

\paragraph{Continuity metrics.}
Overlap RMSE compares predictions aligned to the same absolute control steps
within the overlap of $O$ control steps. Let $\mathcal C$ contain the non-gripper action
dimensions and $d_c=|\mathcal C|$. Using the action chunks defined in
Section~\ref{sec:async-execution}, we compute
\[
E_{\mathrm{ovlp}}^{k}(S)=
\frac{\|\widehat{\mathbf A}_{t_k}[S:R,\mathcal C]-\widehat{\mathbf A}_{t_{k+1}}[0:O,\mathcal C]\|_F}
{\sqrt{O d_c}}.
\]
Slices are zero-indexed with exclusive upper endpoints. Jerk is a per-step
proxy computed from measured non-gripper joint states $q_t$: we take the RMS
of $q_{t+3}-3q_{t+2}+3q_{t+1}-q_t$ over windows that cross a chunk handoff,
without dividing by $\Delta t^3$. Both metrics are computed per episode and
then averaged equally over episodes with valid measurements, including
successful and failed episodes.

\begin{table}[htbp]
\centering
\caption{\textbf{Effect of asynchronous overlap length.}
With $R=24$, increasing the trigger stride $S$ shortens the overlap $O$.
RMSE and jerk decrease throughout the tested range, whereas the highest observed success rate (SR) occurs
at $S=16$. Stride and overlap are measured in control steps.}
\label{tab:async-overlap}
\small
\begin{tabular*}{0.84\linewidth}{@{\extracolsep{\fill}}rrrrr@{}}
\toprule
$S$ & Overlap $O$ & SR (\%) $\uparrow$ & RMSE $\downarrow$ & Jerk $\downarrow$ \\
\midrule
12 & 12 & 94.2 & 0.0246 & 0.0436 \\
16 & 8 & \textbf{95.0} & 0.0063 & 0.0156 \\
20 & 4 & 94.3 & \textbf{0.0058} & \textbf{0.0143} \\
\bottomrule
\end{tabular*}
\end{table}

\paragraph{Smoothness and decision frequency.}
Both RMSE and jerk decrease as the overlap shortens, with a larger reduction
from 12 to 8 steps than from 8 to 4 steps. This trend motivates the
short-overlap execution enabled by streaming VAE
(Section~\ref{sec:async-execution}). The success rate does not improve
monotonically with smoothness: it rises from
94.2\% to 95.0\%, then falls to 94.3\%. We interpret this pattern as a
trade-off between action continuity and decision frequency. At a fixed
control interval $\Delta t$, increasing $S$ lengthens the nominal interval
between inference decisions, $S\Delta t$, reducing the nominal decision frequency
to $f_{\mathrm{decision}}=1/(S\Delta t)$ when no handoff waits occur.
The robot still executes actions at the control rate $1/\Delta t$.
Slower feedback can therefore offset the benefit of smoother
chunk transitions. Minimizing discontinuity alone need not maximize task success.

\section{IDM versus Co-Denoising: Capability and Inference Latency}
\label{app:idm-codenoise}

Long-WAM (IDM) retains video-first causal imagination: it predicts future
visual latents independently of action tokens, then denoises actions
conditioned on those predictions (Section~\ref{sec:adaptation}). Co-denoising
(CoD) instead updates future video and actions jointly, with bidirectional
interaction at each denoising step. IDM provides the action expert with an explicit
prediction of the future before generating actions, at the cost of sequential video
and action inference. IDM achieves modestly higher reported average success
rates on both LIBERO and RoboTwin~2.0
(Tables~\ref{tab:libero} and~\ref{tab:robotwin}).

\paragraph{Latency protocol.}
We compare optimized end-to-end inference latency on NVIDIA GeForce
RTX~5090, including the full observation VAE computation but excluding
preprocessing, text encoding, and controller/IPC overhead. Both variants
use video NVFP4, BF16 action compute and KV storage, and observed-video KV reuse.
For CoD, the first joint step builds the history KV cache for reuse by later
steps; future-video and action KV continue to be updated jointly. Each variant uses
its own trained denoising schedule.

Reported latencies are the mean of the medians from two
independent processes, each with five excluded warmups and 30 steady-state samples.
The cumulative V4/A4 study in Table~\ref{tab:efficiency-cumulative-ablation}
uses the same aggregation and sample counts, excludes cold stabilization,
and includes the full observation VAE and inference computation while excluding
setup, text encoding, controller overhead, and IPC.
Table~\ref{tab:idm-codenoise-latency} combines the IDM measurements from
Table~\ref{tab:efficiency-model-latency} with the CoD measurements.
These timing runs are separate from the task-success evaluations above;
the comparison uses each variant's trained schedule and cache implementation,
so it does not isolate execution order as the sole source of the latency difference.

\begin{table}[t]
\centering
\caption{\textbf{Optimized IDM and CoD inference on RTX~5090.}
V4/A4 and V2/A2 use four and two steps per expert, respectively; CoD shares
the denoising schedule across experts. Latency includes observation VAE computation.
Rates are the reciprocals of inference latencies, not robot control frequencies.}
\label{tab:idm-codenoise-latency}
\small
\begin{tabular*}{0.92\linewidth}{@{\extracolsep{\fill}}lrrrr@{}}
\toprule
& \multicolumn{2}{c}{V4/A4} & \multicolumn{2}{c}{V2/A2} \\
\cmidrule(lr){2-3}\cmidrule(l){4-5}
Method & Latency (ms) $\downarrow$ & Rate (Hz) $\uparrow$ & Latency (ms) $\downarrow$ & Rate (Hz) $\uparrow$ \\
\midrule
Long-WAM (IDM) & 107.4 & 9.3 & 81.8 & 12.2 \\
Long-WAM (CoD) & 90.9 & 11.0 & 65.7 & 15.2 \\
\bottomrule
\end{tabular*}
\end{table}

\paragraph{Capability and compute rate.}
CoD takes 90.9\,ms with V4/A4 and 65.7\,ms with V2/A2, compared with
107.4 and 81.8\,ms for IDM (Table~\ref{tab:idm-codenoise-latency}).
IDM achieves higher success in the separate task evaluations, while its
optimized inference reaches 9.3 and 12.2\,Hz at the two budgets. The two-step
configuration therefore fits within a 100\,ms inference-compute budget.
Online execution must also accommodate
transfer and scheduling costs (Equation~\ref{eq:async-deadline}); asynchronous
execution overlaps inference with ongoing robot motion. The real-world results
in Section~\ref{sec:real-world} provide separate evidence that IDM's video-first
design supports responsive manipulation.

\paragraph{Observed-video KV reuse in IDM.}
Observed-video KV reuse changes token shapes. On RTX~5090, shape-specific
dispatch extends the base quantizer's 588-token tile choice to the 392/196-token
inputs produced by reuse. With the complete V4/A4 configuration, latency is 107.4\,ms with reuse versus 112.6\,ms without it (4.6\%).
Spark reaches 328.2\,ms versus 394.9\,ms without reuse (16.9\%).
With V2/A2, reuse gives no latency reduction on RTX~5090
(81.8 versus 81.3\,ms), while latency on Spark decreases from 280.8 to
254.4\,ms (9.4\%). These comparisons hold the remaining configuration fixed
within each device and denoising budget. They differ from the reuse-stage gains
in Table~\ref{tab:efficiency-cumulative-ablation}, which are measured before
subsequent device-specific tuning.

\section{Context-Dependent Inference Latency}
\label{app:context-latency}

Table~\ref{tab:context-latency-5090} shows efficient long-context inference
with our infrastructure on RTX~5090: increasing history eightfold, from
2.4 to 19.2 seconds, raises end-to-end chunk latency from 107.4 to
341.0\,ms (approximately $3.2\times$). These results use the shared optimizations and device-specific tuning
in Section~\ref{sec:edge-acceleration}. For
online execution, streaming VAE additionally moves prefix encoding ahead
of the inference trigger, reducing post-trigger preparation work
(Section~\ref{sec:async-execution}).

\begin{table}[t]
\centering
\caption{\textbf{Context-dependent latency on RTX~5090.}
End-to-end inference time per action chunk with the Long-WAM infrastructure.
Context $P$ counts preceding control intervals, corresponding to $P/20$
seconds of history; $P=0$ retains the current observation.}
\label{tab:context-latency-5090}
\small
\setlength{\tabcolsep}{10pt}
\renewcommand{\arraystretch}{1.08}
\begin{tabular}{rrr}
\toprule
Context $P$ & History (s) & Latency (ms) $\downarrow$ \\
\midrule
  0 &  0.0 &  74.6 \\
 48 &  2.4 & 107.4 \\
 96 &  4.8 & 138.3 \\
192 &  9.6 & 204.5 \\
384 & 19.2 & 341.0 \\
\bottomrule
\end{tabular}
\end{table}

\section{Additional Real-World Deployment Visualizations}
\label{app:real-world-details}

We provide detailed rollouts complementing the real-robot evaluation in
Section~\ref{sec:real-world}. The montages retain the original overview and
wrist-camera views, with red annotations marking failure events and green
annotations marking successful outcomes. These selected examples illustrate
execution behavior; aggregate success rates are reported in the main text.

\subsection{Dynamic Composite Manipulation}
\label{app:real-composite}

\begin{figure}[!t]
\centering
\includegraphics[width=\linewidth]{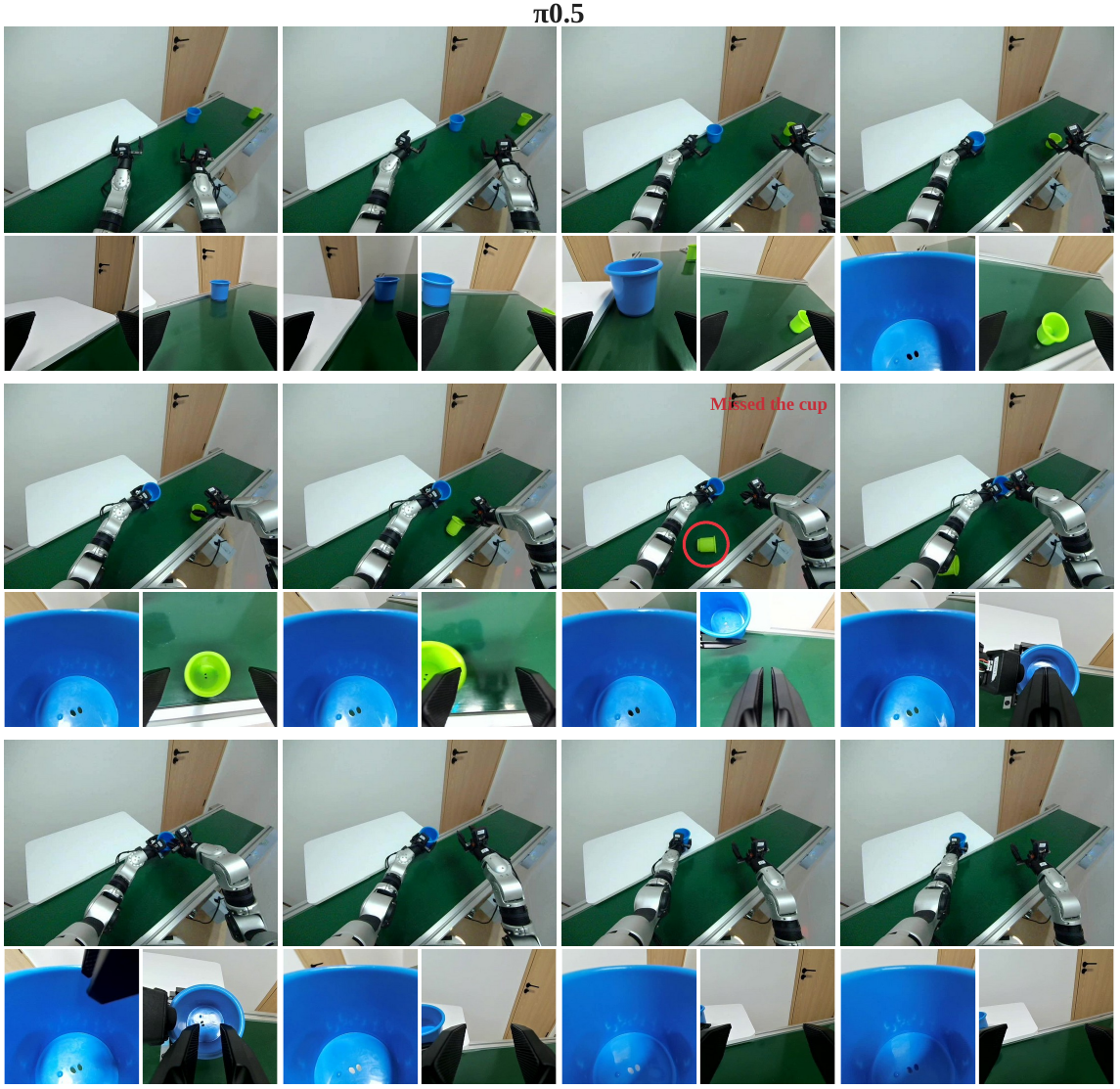}
\caption{\textbf{Dynamic cup stacking with $\pi_{0.5}$.}
Time progresses left to right within each four-column block, then continues
in the next block below; each overview frame is accompanied by two wrist
views. The policy grasps the blue cup but misses the moving green cup
(red circle), leaving the composite task incomplete.}
\label{fig:app-composite-pi05}
\end{figure}

\begin{figure}[!t]
\centering
\includegraphics[width=\linewidth]{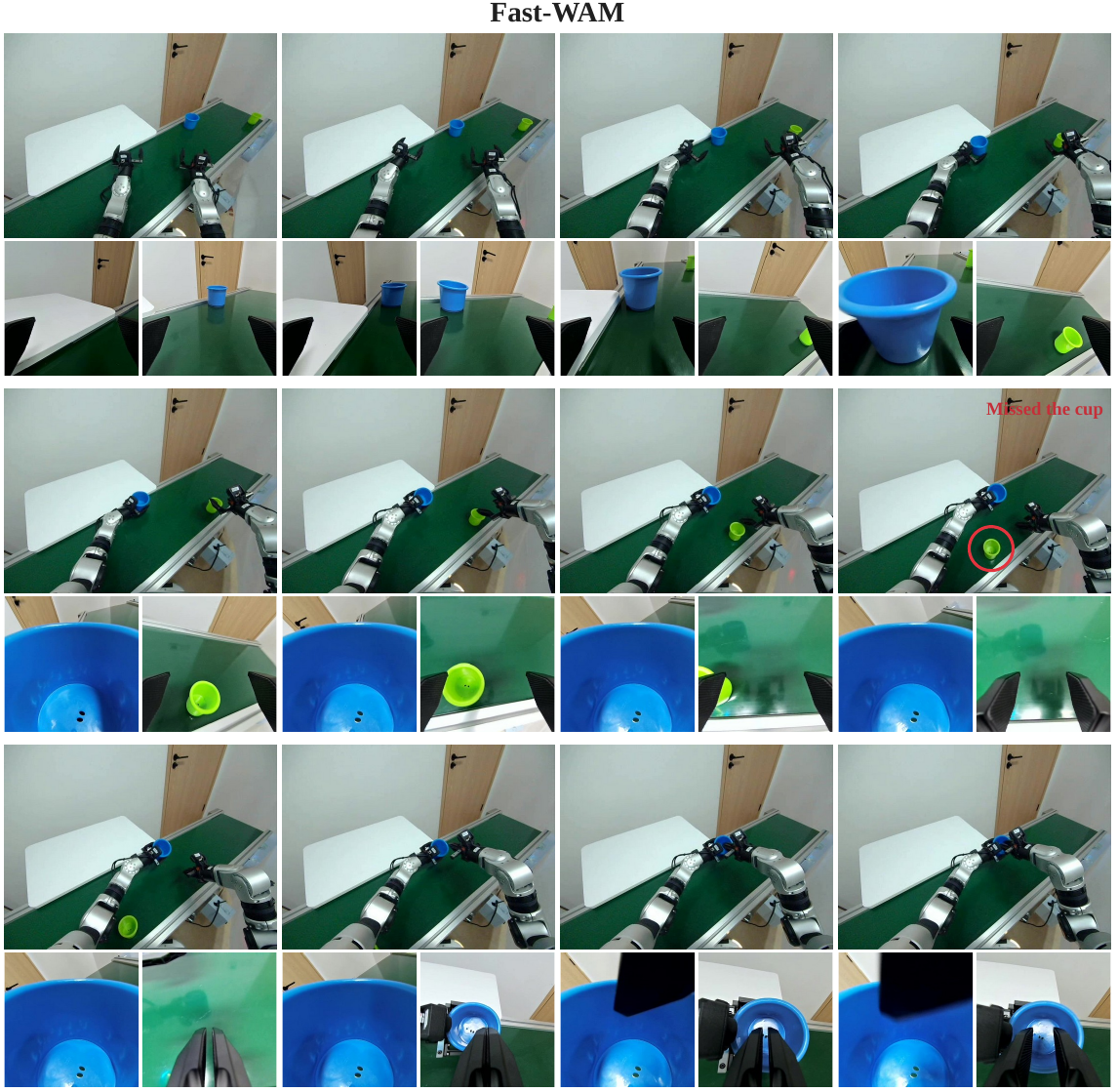}
\caption{\textbf{Dynamic cup stacking with Fast-WAM.}
The sequence follows the same reading order as
Figure~\ref{fig:app-composite-pi05}. After grasping the blue cup, the policy
reaches toward the green cup but misses it as it moves along the conveyor.
The subsequent frames show that the nesting stage is not completed.}
\label{fig:app-composite-fastwam}
\end{figure}

\begin{figure}[!t]
\centering
\includegraphics[width=\linewidth]{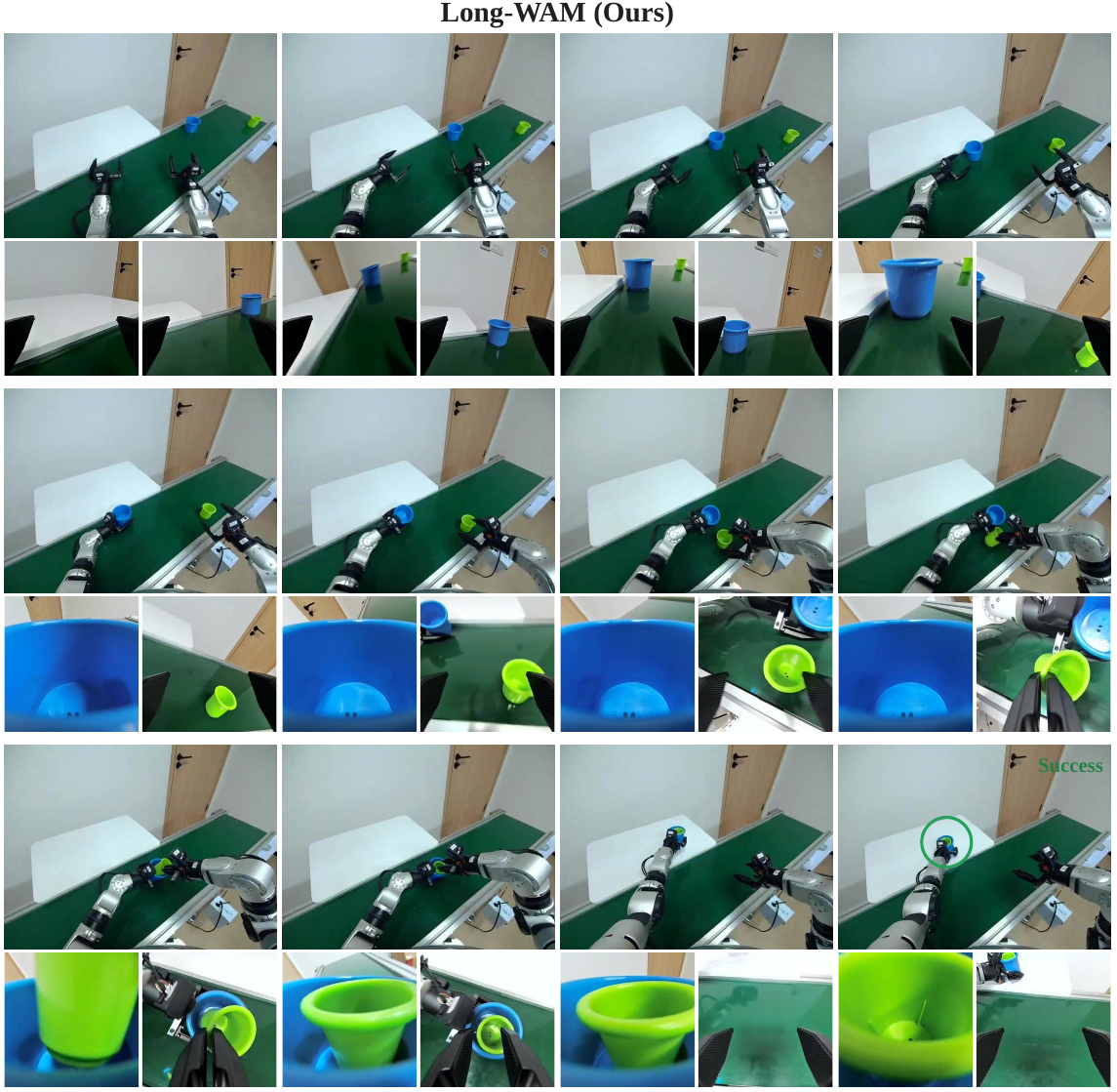}
\caption{\textbf{Dynamic cup stacking with Long-WAM.}
Long-WAM grasps the blue cup, intercepts the moving green cup, and places
the green cup inside the blue cup. The overview and wrist views reveal
the transition from interception to alignment and insertion, illustrating
coordinated execution across the stages of this dynamic task.}
\label{fig:app-composite-longwam}
\end{figure}

Figures~\ref{fig:app-composite-pi05}--\ref{fig:app-composite-longwam}
show dynamic cup stacking on Unitree G1. Long-WAM coordinates grasping the
blue cup, intercepting the moving green cup, and nesting it inside the blue
cup. The baseline rollouts instead miss the green cup during interception.

\subsection{Moving-Object Grasping at Different Speeds}
\label{app:real-speed}

\begin{figure}[!t]
\centering
\includegraphics[width=\linewidth,height=0.77\textheight,keepaspectratio]{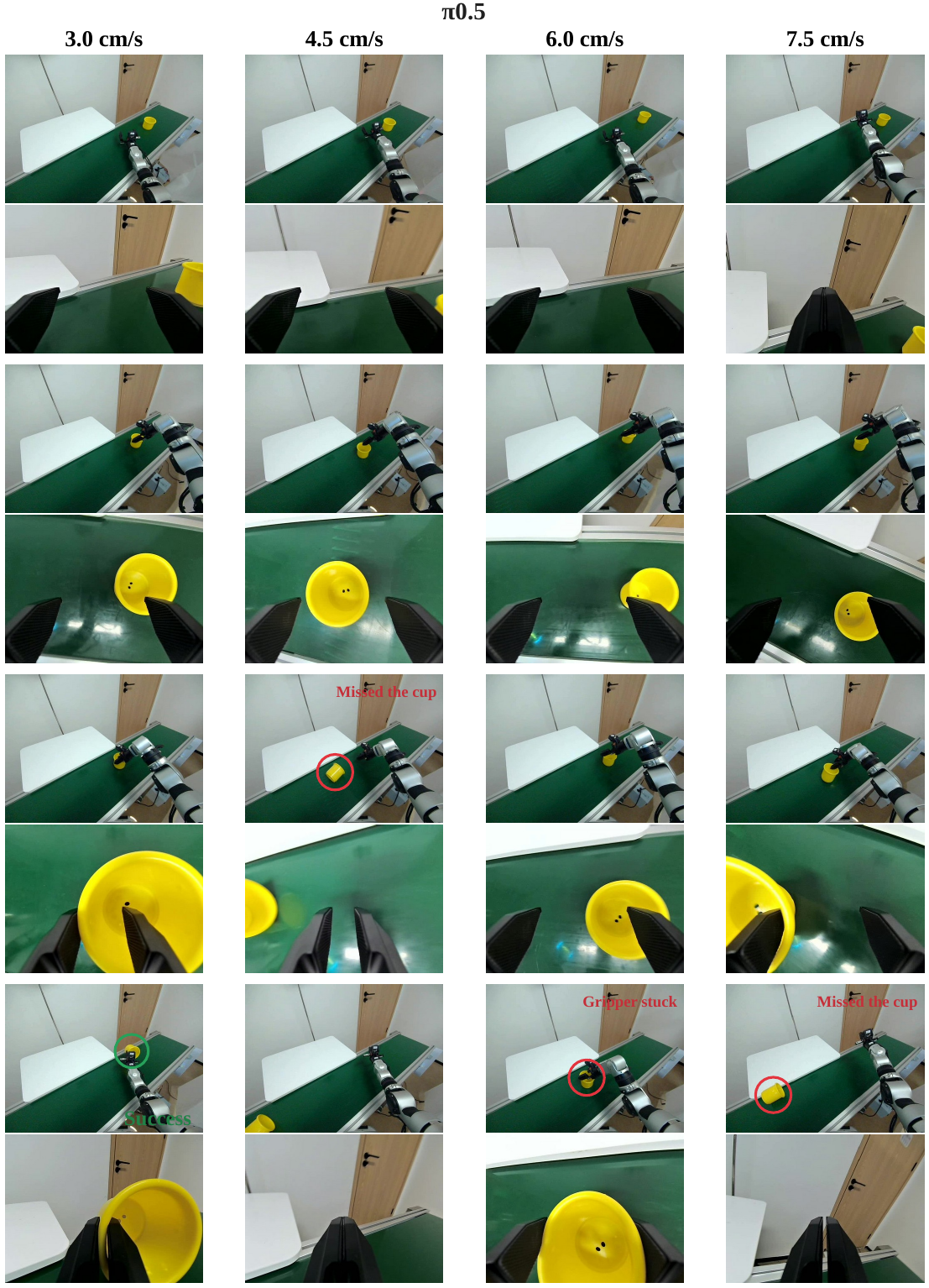}
\caption{\textbf{Moving-object grasping with $\pi_{0.5}$.}
Columns show 3.0, 4.5, 6.0, and 7.5~cm/s; time advances downward through
paired overview and wrist views. The selected rollout succeeds at 3.0~cm/s,
misses the cup at 4.5 and 7.5~cm/s, and exhibits the annotated gripper-stuck
failure at 6.0~cm/s.}
\label{fig:app-speed-pi05}
\end{figure}

\begin{figure}[!t]
\centering
\includegraphics[width=\linewidth,height=0.77\textheight,keepaspectratio]{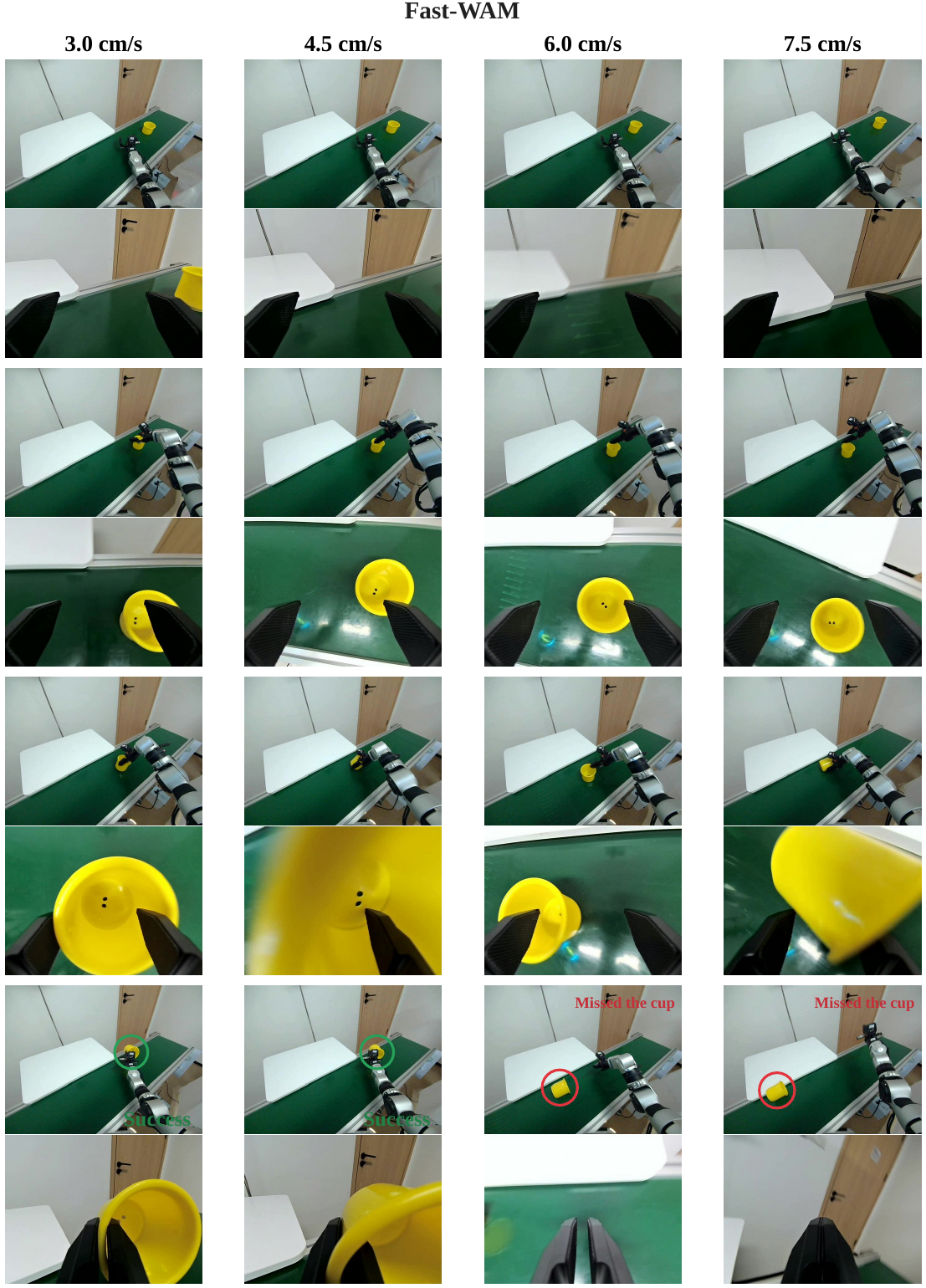}
\caption{\textbf{Moving-object grasping with Fast-WAM.}
Columns and temporal ordering match Figure~\ref{fig:app-speed-pi05}.
The selected rollouts succeed at 3.0 and 4.5~cm/s but miss the cup at
6.0 and 7.5~cm/s. The wrist views show the target moving beyond the
gripper before a secure grasp is established.}
\label{fig:app-speed-fastwam}
\end{figure}

\begin{figure}[!t]
\centering
\includegraphics[width=\linewidth,height=0.77\textheight,keepaspectratio]{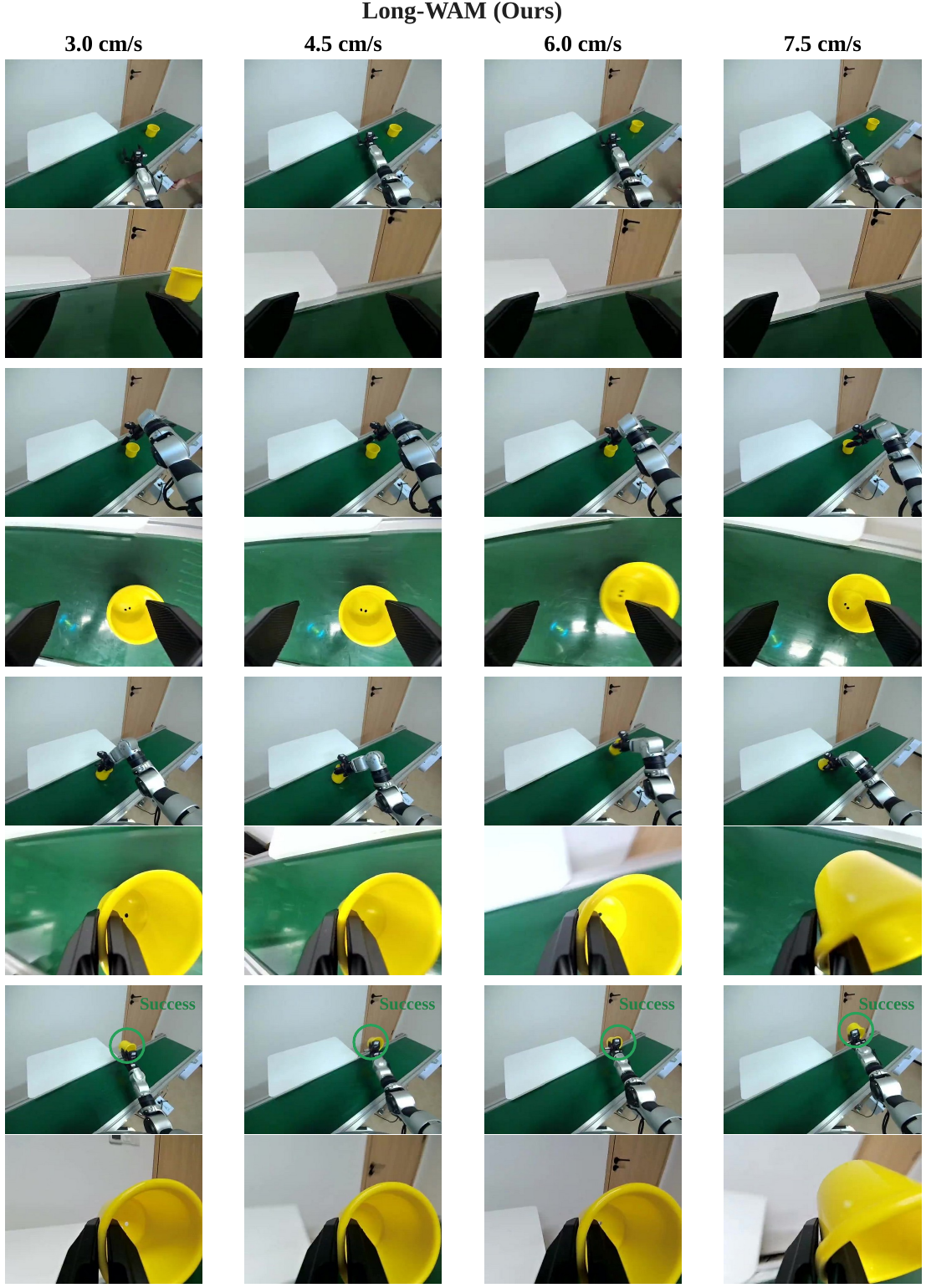}
\caption{\textbf{Moving-object grasping with Long-WAM.}
The illustrated rollouts complete the grasp at all four conveyor speeds,
including 6.0 and 7.5~cm/s. Successive wrist views show the moving cup
entering the gripper and being retained after closure, illustrating
responsive interception under progressively tighter timing constraints.}
\label{fig:app-speed-longwam}
\end{figure}

Figures~\ref{fig:app-speed-pi05}--\ref{fig:app-speed-longwam} compare conveyor
grasping at four speeds. The baseline examples expose missed interceptions
at higher speeds, whereas Long-WAM completes the illustrated grasp at every
speed, consistent with the dynamic-control results in
Section~\ref{sec:real-dynamic-speed}.

\subsection{Long-Horizon Manipulation}
\label{app:real-long-horizon}

\begin{figure}[!ht]
\centering
\includegraphics[width=\linewidth]{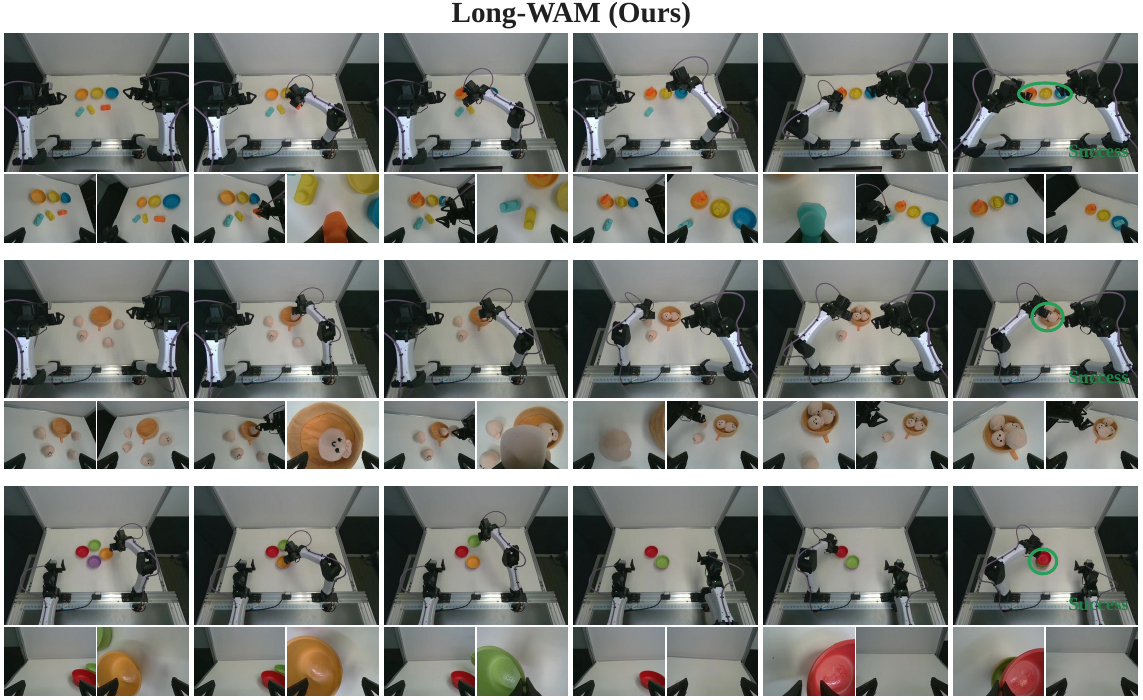}
\caption{\textbf{Long-horizon manipulation with Long-WAM on YAM.}
Top to bottom: brick sorting by color, placing dumplings in a pan, and
stacking bowls. Each task contains six chronological snapshots, read left
to right, with paired wrist views below each overview frame. The final
snapshots show successful task configurations across all three tasks.}
\label{fig:app-long-horizon}
\end{figure}

Figure~\ref{fig:app-long-horizon} shows Long-WAM executing three YAM tasks
that require successive object interactions. The sequences illustrate
task-directed progress through repeated pickup and placement, complementing
the fast dynamic behaviors above. These tasks last over 40 seconds on
average (Section~\ref{sec:real-long-horizon}).

\clearpage

\end{document}